\documentclass{article}
\usepackage[preprint]{neurips_2019}
     
\usepackage{Definitions}
\usepackage{algorithmic}
\usepackage{algorithm}
\usepackage{enumitem}
\usepackage[utf8]{inputenc} 
\usepackage[T1]{fontenc}    
\usepackage{hyperref}       
\usepackage{url}            
\usepackage{booktabs}       
\usepackage{amsfonts}       
\usepackage{nicefrac}       
\usepackage{microtype}      
\usepackage{mathtools}
\usepackage{subcaption}
\usepackage{color}

\theoremstyle{theorem}
\newtheorem{theorem}{Theorem}[section]
\newtheorem{lemma}[theorem]{Lemma}

\theoremstyle{definition}
\newtheorem{definition}[theorem]{Definition}
\theoremstyle{corollary}
\newtheorem{corollary}[theorem]{Corollary}
\theoremstyle{proposition}
\newtheorem{proposition}[theorem]{Proposition}

\newtheorem{remark}[theorem]{Remark}

\numberwithin{equation}{section}

\newcommand{\proj}[2]{\operatorname{\Pi}_{#1}^{#2}}
\newcommand{\grad}{\nabla}

\newcommand{\masteralg}{Master Gradient Descent }
\newcommand{\fastmasteralg}{Fast Master Gradient Descent }
\newcommand{\masterfrm}{Master OCO Framework }
\newcommand{\ogdlike}{Gradient-Based }

\newcommand{\simplex}{\Delta}
\newcommand{\kl}[2]{\mathrm{KL}(#1\,\|\,#2)}
\newcommand{\MD}{\mathsf{OMD}}
\newcommand{\OCO}{\mathsf{OCO}}

\renewcommand{\S}{Section}
\newcommand{\Apn}{Appendix}

\title{Adaptive Online Learning for Gradient-Based Optimizers}
\author{%
		Saeed Masoudian \quad Ali Arabzadeh  \quad  Mahdi Jafari Siavoshani\\
		Computer Engineering Departement\\
		Sharif University of Technology\\
		  Tehran, Iran \\
		\texttt{\{masoodian, arabzadeh\}@ce.sharif.edu, mjafari@sharif.edu} \\
		\And
		Milad Jalali\\
		Computer Engineering Departement\\
		Sharif University of Technology\\
		Tehran, Iran \\
		\texttt{jalali@ce.sharif.edu} \\
		\And
		Alireza Amouzad\\    
		Computer Engineering Departement\\
		Amirkabir University of Technology\\
		  Tehran, Iran \\
		\texttt{a.r.amouzad.m@gmail.com} \\
}

\begin{document}
		
		\maketitle

		\begin{abstract}
As application demands for online convex optimization accelerate, the need for designing new methods that simultaneously cover a large class of convex functions and impose the lowest possible regret is highly rising. 
Known online optimization methods usually perform well only in specific settings, and their performance depends highly on the geometry of the decision space and cost functions.
However, in practice, lack of such geometric information leads to confusion in using the appropriate algorithm. To address this issue, some adaptive methods have been proposed that focus on adaptively learning parameters such as step size, Lipschitz constant, and strong convexity coefficient, or on specific parametric families such as quadratic regularizers. In this work, we generalize these methods and propose a framework that competes with the best algorithm in a family of expert algorithms. Our framework includes many of the well-known adaptive methods including MetaGrad, MetaGrad+C, and Ader. 
	We also introduce a second algorithm that computationally outperforms our first algorithm with at most a constant factor increase in regret. Finally, as a representative application of our proposed algorithm, we study the problem of learning the best regularizer from a family of regularizers for Online Mirror Descent. 
	Empirically, we support our theoretical findings in the problem of learning the best regularizer on the simplex and $l_2$-ball in a multiclass learning problem.
\end{abstract}
		\section{Introduction}\label{intro}

Online Convex Optimization (OCO) plays a pivotal role in modeling various real-world learning problems such as prediction with expert advice, online spam filtering, matrix completion, recommender systems on data streams and large-scale data \cite{Intro:Online:Convex}. The formal setting an OCO is described as follows.
\paragraph{OCO Setting}\label{OCO} In OCO problem~\cite{bianchi-2006-prediction, Intro:Online:Convex, Online:suvery},  at each round $t$, we play $x_t \in \mathcal{D}$ where $\mathcal{D} \subseteq \mathbb{R}^d$ is a convex set. The adversarial environment incurs a cost $f_t(x_t)$ where  $f_t(x)$ is a convex cost function on $\mathcal{D}$ at iteration $t$. 
The main goal of OCO is to minimize the cumulative loss of our decisions. Since losses can be chosen adversarially by the environment, we use the notion of Regret as the performance metric, which is defined as 
\begin{equation}
\mathcal{R}_T \triangleq \sum\limits_{t=1}^T f_t(x_t) - \min\limits_{x \in \mathcal{D}} \sum\limits_{t=1}^T f_t(x).
\end{equation}

In fact, \emph{regret} measures the difference between the cumulative loss of our decisions and the best static decision in hindsight. In the literature, various iterative algorithms for OCO problem try to  minimize regret and provide sublinear upper bound on it. All these algorithms are variations of \emph{Online Gradient Descent} (OGD); meaning that these algorithms share a common feature in their \emph{update rule} \cite{Intro:Online:Convex, Online:suvery, JMLR:Adaptive}. Furthermore, their updating process is performed just based on previous decision points and their gradients. We call this family of OCO as \ogdlike algorithms. In this paper, our attention is mainly drawn to this family of algorithms. Some of these algorithms such as \emph{Online Newton Steps} \cite{ML:Hazan:2007} and \emph{AdaGrad} \cite{JMLR:Adaptive, McMahanStreeter2010} have considered a specific class of cost functions like strongly-convex, exp-concave, and smooth functions. Then, by manipulating the step size and using second-order methods, they have been able to reach a better regret bound than $\mathcal{O}(\sqrt{T})$ \cite{ML:Hazan:2007}. If we have no other restriction than convexity on cost functions, then the \emph{Regularization} based algorithms such as \emph{Follow The Regularized Leader (FTRL)} \cite[page 72]{Intro:Online:Convex} and \emph{Online Mirror Descent}  \cite[page 76]{Intro:Online:Convex} step into the field. In these algorithms, the geometry of the domain space $\mathcal{D}$ has been taken into account and in spite of the fact that their regret's upper bound remains $\mathcal{O}(\sqrt{T})$, the constant factor of their regret bound can be improved by choosing a suitable regularizer.

Each \ogdlike algorithm that performs on Lipschitz functions has the regret upper-bound $\mathcal{O}(\sqrt{T})$ and based on \cite[page 45]{Intro:Online:Convex} this bound is tight ($\ie$  for each algorithm there is a sequence of cost functions whose regret is $\Omega(\sqrt{T})$). However, the constant factor in these algorithms is different. 

In summary, there exists a group of iterative algorithms each of them has a number of tuning parameters. Consequently, in OCO setting it is very important to choose the right algorithm with the best set of parameters such that it results to the lowest regret bound w.r.t. the geometry of space and choice of cost functions. However, due to lack of our knowledge about the problem setup, it is not always possible to choose the right algorithm or tuning parameters.
Our aim is to introduce a \emph{master} algorithm that can compete with the best of such iterative algorithms in terms of regret bound.

 \subsection{Related Works}
 It is known that OGD achieves $\mathcal{O}(\sqrt{T})$ regret bound \cite[page 43]{Intro:Online:Convex} . In addition, if cost functions are strongly convex, then the regret bound  $\mathcal{O}(\log{T})$ can be achieved \cite{Online:suvery}. It is shown that \emph{Online Newton Step} for exponentially concave cost functions has $\mathcal{O}(d\log{T})$ regret bound \cite{ML:Hazan:2007}. 
 
 Considering adaptive frameworks, numerous approaches have been proposed in the literature to learn the parameters of OGD algorithm like \emph{step-size} \cite{van2016metagrad} and \emph{diameter} of $\mathcal{D}$ \cite{cutkosky2017online}. For tuning regularizer, one can mention \emph{AdaGrad} algorithm that learns from a family of Quadratic Matrix regularizers \cite{JMLR:Adaptive}. 
 AdaGrad is a special case of the work presented in \cite{McMahanStreeter2010} that uses a family of increasing regularizers.
 \emph{MetaGrad} algorithm that was proposed later than  \emph{AdaGrad} in \cite{van2016metagrad} has the ability to learn the \emph{step-size} for all Gradient-Based algorithms. However, it has high time complexity and needs many oracle accesses per iteration.

		\section{Preliminaries}\label{sec:preliminaries}
In this section, we will introduce the notation that will be used throughout the paper and review some of the preliminary materials required to introduce our method.
\subsection{Notation}
We keep the following notation throughout the rest of paper.
We use $V_{1:n}$ to denote a sequence of vectors $(V_1, \ldots, V_n)$.
Let $x_t$ and $f_t$ be our decision and cost function respectively, then $\grad_t$ denotes $\grad f_t(x_t)$. 
For cost function $f_t(x)$, surrogate cost function is denoted by $\widehat{f}_t(x) = \inner{\grad_t}{x}$. 
Denote the upper bound on surrogate cost functions by $F = \sup\limits_{t \in \NN , x \in \Dcal} | \inner{\grad f_t(x)}{x} |$.
Projection of vector $y$ on domain $\Dcal$ w.r.t. some function or norm $R$ is denoted by $\proj{\Dcal}{R}(y)$.
Moreover, we denote by $\BB_p$ the unit ball in $\RR^d$ for $\ell_p$ norm, \ie, $\BB_p^d = \{x\in\RR^d \mid \|x\|_p \leq 1\}$.
Also, we denote $\simplex(n)$ to be the $n$-simplex, \ie,
$\simplex(n) = \{ x\in\RR_+^n \mid \one^\top x = 1\}$.
Finally, each OCO algorithm has its own regret bound on a family of cost functions. 
To refer to the regret bound of an arbitrary algorithm $\Acal$ after $T$ iterations, we use the notation $\Bcal_T^\Acal$.

\begin{definition}
As mentioned in \S~\ref{intro}, \ogdlike algorithms are algorithms whose \emph{update rule} is performed just based on previous decision points and their gradients. So for an arbitrary \ogdlike algorithm \emph{A}, we have an iterative update rule
$x_{t} = \Psi_{A}(x_{t-1}, \grad_{1:t-1})$
and a non-iterative or closed form update rule denoted by
$x_t = \Upsilon_{A}(x_0, \grad_{1:t-1})$.
\end{definition}

In general, it can be difficult to  derive the closed form for an algorithm. However, for some algorithms like OGD, Online Mirror Descent (OMD), AdaGrad, \etc, $\Upsilon$ can be efficiently computed and eventually, attain the same complexity as $\Psi$. In Proposition \ref{prop:ComputeUpdateRuleMD}, we show how to efficiently compute the update rules of OMD and AdaGrad.

\subsection{Problem Statement}\label{sec:ProbState}

In this work, we focus on learning the best algorithm among a family of OCO algorithms. We also define the problem of learning the best regularizer as a special yet important case of learning the best OCO algorithm. Both problems are explicitly defined in the following.

	\textbf{Best OCO Algorithm:} Let $\mathcal{D} \in \mathbb{R}^d$ be a compact convex set that presents the search domain of an OCO Algorithm. Our focus is on \ogdlike algorithms, so we have a family $\mathcal{M} \triangleq \{\OCO_1, \ldots, \OCO_k\}$ of algorithms where the  update rule of the $i$-th algorithm is given by $x_{t+1} = \Psi_i(x_t, \grad_{1:t})$. Our goal is  to propose an algorithm that perform as good as the best algorithm in $\Mcal$.

\textbf{Best Regularizer:} When the family of algorithms  only contains OMD algorithms, each member of $\mathcal{M}$ is completely characterized by its \emph{Regularizer}. We consider $\MD_\varphi$ as an OMD algorithm with Regularizer $\varphi(x)$. Now, let $\Phi \triangleq \{\varphi_1(x), \ldots, \varphi_k(x)\}$ be the set of regularizers in which the $i$-th element is $\eta_i$ strongly convex w.r.t. a norm $\|\cdot\|_i$. So we have a set of OMD algorithms with regularizers $\Phi$ denoted by $\mathcal{M} = \{\MD_{\varphi_1}, \ldots, \MD_{\varphi_k}\}$. Moreover, we have an OCO problem similar to the ``best OCO algorithm'' defined above,
that at each iteration decides based on the performance of all OMD algorithms in $\mathcal{M}$ (more precisely, best of them).

\subsection{Expert Advice} \label{expert_advice}
 Suppose we have access to $k$ experts $a_1, \ldots, a_k$. At each round $t$, we want to decide based on the decisions of experts and then incur some loss $\ell_t(a_t) \in [0,1]$ from the environment as feedback. This problem can be cast into the  \emph{Online Learning} in which to evaluate the goodness of an
 algorithm, the notion of Regret is used. Here, we use	$\mathcal{R}_T(a^*) = \sum_{t=1}^T \ell_t(a_t) - \sum_{t=1}^T \ell_t(a^*)$
to denote the regret of expert $a^*$.
 All algorithms for expert advice problem, follow the  iterative framework described below \cite{bianchi-2006-prediction,Freund:1997,vas-90,bubeck12regret,Vovk1998,LittlestoneWarmuth1994} . 
\paragraph{Expert Advice Framework}  Let $p_t$ be the probability of choosing experts in each iteration.  Suppose that based on prior knowledge we have a distribution $p_1$ over experts. If we have no idea about the experts, $p_1$ can be chosen to have a uniform distribution.  At  iteration $t$, we choose expert $a_t \propto p_t$ and play the decision made by $a_t$. Then the loss vector $\ell_t$ can be observed. We will update the probabilites $p_{t+1}$ based on losses we have observed until now. 



 In the expert advice framework, we can have two different settings based on the availability of feedbacks, stated as follows. 
(1) Full feedback setting where all experts losses $\ell_t$ are observed.  
(2) Limited feedback setting,  or the so called \emph{Bandit} \cite{bubeck12regret} version, where only  $\ell_t(a_t)$ is observed.

In what follows, the regret bounds of two well known algorithms namely \emph{Hedge} \cite{Freund:1997}  and \emph{Squint} are explained. We will elaborate on \emph{exponential-weight algorithm for exploration and exploitation (EXP3)} \cite{auer2002nonstochastic} and \emph{gradient based prediction algorithm (GBPA)} \cite{abernethy2015fighting} in the bandit setting. 
 
\begin{theorem}[\cite{Freund:1997}]
Hedge algorithm, defined by choosing $p_t(a) \propto \exp\left( -\eta \sum\limits_{\tau=1}^{t-1} \ell_\tau(a)\right)$ in the expert advice framework, ensures  ~
$
	\EE\left(\mathcal{R}_T(a^*)\right) \leq \frac{\log(K)}{\eta} + \eta T \leq 2\sqrt{\log(K)T}.
$
\end{theorem} 

\begin{theorem}[\cite{squint}]\label{theorem:squint}
Let $r_t(i) = \inner{p_t}{\ell_t} - \ell_t(i)$ and $V_T(a) = \sum\limits_{t=1}^{T}  r_t(a)^2$. Then the Squint algorithm, defined by  $p_t(a) \propto p_1(a) \exp\left( -\eta \sum\limits_{\tau=1}^{t-1} r_\tau(a) + \eta^2\sum\limits_{\tau=1}^{t-1} r_\tau(a)^2 \right)$ in 
the expert advice framework, ensures
$	\EE(\mathcal{R}_T(a^*)) \leq \frac{ln(1/p_1(a^*)}{\eta}+ \eta V_T(a^*)	\leq  2\sqrt{ V_T(a^*) ln(1/p_1(a^*))}$.
\end{theorem} 
\begin{theorem}[\cite{abernethy2015fighting}]\label{thm:GBPA_RegretBnd}
GBPA algorithm, uses estimated loss $\widehat{\ell}_t = \frac{\ell_t(a_t)}{p_t(a_t)}\bold{e}_{a_t}$  and update rule $p_t(a) \propto  \grad (\eta S_\alpha)^*\big( -\eta \sum\limits_{r=1}^{t-1} \widehat{\ell}_r(a)\big)$ in the expert advice framework, where $S_\alpha$ is Tsallis entropy with parameter $\alpha$, ensures
$
	\EE(\mathcal{R}_T(a^*)) \leq \eta \frac{K^{1-\alpha}-1}{1-\alpha} + \frac{K^\alpha T}{2\eta \alpha}  \leq 	4\sqrt{KT}
$ where 
$\alpha$ chooses as $1/2$.
\end{theorem}

\begin{corollary}\label{corollary:exp3}
	In Theorem~\ref{thm:GBPA_RegretBnd}, if $\alpha \rightarrow 1$ leads to $p_t(a) \propto \exp\big( -\frac{1}{ \eta} \sum\limits_{r=1}^{t-1} \widehat{\ell}_r(a)\big)$. So EXP3 algorithm is recovered and ensures ~$
	\EE(\mathcal{R}_T(a^*)) \leq \eta \log(K) + \frac{TK}{2\eta } \leq  \sqrt{2K\log(K)T}
$.
\end{corollary}


\begin{remark}
	If we know that $\forall i,t : \ell_t(i) < L$, then in all expert advice theorems, the regret bounds will be multiplied by a factor $L$.
\end{remark}

%
%
%
%
%
%

\subsection{Online Mirror Descent}
	\begin{definition}[Online Mirror Descent]\label{def:MD}
	Update rule $\Psi$ for \emph{lazy} and \emph{agile} versions of OMD with regularizer $\varphi$ are defined as 
	\begin{equation}\label{eq:md_update}
\begin{aligned}
Agile: & \hspace{.5cm} x_{t+1} = \Psi(x_t, \grad_{1:t}) = \proj{\Dcal}{\varphi}( \grad \varphi^*(x_t - \eta \grad_t)),\\
Lazy: & \hspace{.5cm} y_{t+1} = \Psi(y_t, \grad_{1:t}) = \grad \varphi^*(y_t - \eta \grad_t),  x_{t+1} = \proj{\Dcal}{\varphi}(y_{t+1}).
\end{aligned}
\end{equation}
\end{definition}

\begin{proposition}\label{prop:ComputeUpdateRuleMD}
Computing the closed form of $x_t$ for agile version of OMD is very complicated but for lazy update rule, we have $
y_{t+1} = \Upsilon(y_t,\grad_{1:t}) = \proj{\Dcal}{\varphi}\big(\grad \varphi^*(x_0-\eta\sum_{i=1}^t \grad_i)\big)
$.
Thus, the computation of $\Upsilon$ is  light weighted because we need only to keep $S_t = \sum_{i=1}^t \grad_i$ in each iteration.
\end{proposition}
		\section{Proposed Methods}\label{proposed}

Our proposed methods for the problem stated in \S~\ref{sec:ProbState},  are inspired by \emph{expert advice} problem. First, we propose  an algorithm that uses expert advice in full feedback setting and then for the sake of time complexity, present another algorithm that has almost the same regret as the former algorithm

\subsection{Assumptions}\label{sec:assumptions}
Here, we review three assumptions in this work.
(1) All cost functions are Lipschitz w.r.t some norm  $\|x\|_.$  on $\mathcal{D}$, \ie, there exists $L_{.} > 0$ such that
$\forall x, y \in \mathcal{D} :  |f(x)-f(y)| \leq L_{.}\|x-y\|_.$.  
(2) Domain $\mathcal{D}$ contains the origin (if not can be translated) and  is bounded w.r.t. some norm $\|x\|_{.}$, \ie, there exists $D_. > 0$ such that
$ \forall x,y \in \mathcal{D}: \|x-y\|_{.} < D$.
(3)  Suppose $\Acal$ is an arbitrary OCO algorithm that performs on $L$-Lipschitz cost functions, w.r.t. an arbitrary norm $\|\cdot\|_.$, and domains with diameter $D$, w.r.t. the same norm. Then, there exists a \emph{tight} upper bound $\Bcal_T^\Acal$ on the regret $\Rcal_T^\Acal$ that $\Acal$  achieve this bound. Hence $\Bcal_T^{\Acal}$ depends on the parameters $L$, $D$, and $T$.



\subsection{\masterfrm}\label{master_framework}
By the problem setting described in \S~\ref{sec:ProbState}, we have $K$ experts and each of these  experts is a \ogdlike algorithm. 
In order to learn the Best OCO Algorithm, we will take advantage of  \emph{expert advice} algorithms.

\textbf{Framework Overview:}
In our proposed framework, called \masterfrm, we consider an expert advice algorithm $\Acal$ and a family of online optimizers. We want to exploit the expert advice algorithm to track the best optimizer in hindsight. In each round, $\Acal$ selects  an optimizer $a_t$  to see its prediction $x_t^{a_t}$.  Environment reveals cost function $f_t(x)$. Then  we pass the surrogate cost function $\widehat{f}_t(x) = \inner{\grad_t}{x}$ to all optimizers instead of the original cost function. Hence, to be consistent with the expert advice scenario assumptions,  we consider normalized surrogate cost function for losses. So we'll have $ \ell_t(i) = \widehat{f}_t(x_t^i)/(2F) +1/2$, where $F$ is an upper bound for surrogate functions.  Now, based on full or partial feedback assumption of $\Acal$,  we pass $\{\ell_t(i)\}_{i \in [K]}$ or $\ell_t(a_t)$ to $\Acal$, respectively. Finally, $\Acal$ updates probability distribution $p_t$ over experts based on the observed losses.

\begin{remark}
	The main reason why we use surrogate function in place of the original cost function is as follows. 
	Considering the $i$-th expert, using surrogate function leads to generating a sequence of decisions $\{x_t^i\}_{t\in [T]}$.
	This is just similar to the situation where we merely use the $i$-th expert algorithm on an OCO problem whose cost functions at iteration $t$ are $\hat{f_t}(x)$. We will prove this claim in \Apn~\ref{appendix:analysis}.
\end{remark}
 In the following the formal description of our framework  is provided.
 \begin{algorithm}[h]
\floatname{algorithm}{Framework}
\caption {\masterfrm}
\begin{algorithmic}[1]
    \STATE {\textbf{Input:}  Expert advice algorithm $\mathcal{A}$, set of online optimizers $\mathcal{M} = \{\OCO_i\}_{i\in [K] }$ }
    \FOR{t = 1, \ldots, T}
    \STATE {$\Acal$ decides what optimizer $a_t \in [K]$ should be selected}
    \STATE {Ask selected optimizer to get prediction  $x_t^{a_t}$}
     \STATE {Play $x_t=x_t^{a_t}$ and the environment incurs a cost function $f_t(x)$}
      \STATE {Pass the surrogate cost function $\widehat{f}_t(x) = \inner{\grad_t}{x}$ to all optimizers}
      \STATE {Select $S = [K]$ or $S = \{a_t\}$  based on partial or fully feedback property of $\Acal$ } 
     \STATE {Set losses for the observed predictions:  $\forall i \in S :\ell_t(i) = \frac{\widehat{f}_t(x_t^i)}{2F} + \frac{1}{2}$ }
     \STATE {Pass $\{\ell_t^i\}_{i \in S}$ to $\Acal$. Now $\Acal$ can update the probabilities over the experts}
    \ENDFOR
\end{algorithmic}
\label{alg:framework}
\end{algorithm}

\begin{proposition}\label{prop:framework}
Let  $\mathcal{M} = \{\OCO_i\}_{i\in [K] }$ be  \ogdlike optimizers and $\Acal$ be an expert advice algorithm. Then for all $\OCO_i \in \Mcal$, our proposed framework ensures
\begin{equation}\label{eq:prop}
	\Rcal_T \leq 2F \cdot \Rcal_T^{\Acal} + \Rcal_T^{\OCO_i},
\end{equation} 
where $F$ is a tight upper bound for all surrogate cost functions, $\Rcal_T^{\OCO_i}$ is the regret of running $i$-th optimizer on surrogate functions and $\Rcal_T^{\Acal}$ is the general regret of expert advice algorithm $\Acal$.  
\end{proposition}
\begin{remark}
	In general, there is no need to  normalize the cost functions. In fact we can pass surrogate cost functions as losses $\ell_t(i) = \widehat{f}_t(x_t^i)$ and gain the same regret bound as mentioned above. So without knowing $F$, we can still apply the above framework.
\end{remark}

\begin{corollary}
In expert advice algorithm $\Acal$, suppose $p_t$ is the probability distribution over  optimizers at iteration $t$. If we have access to all optimizers' predictions $\{x_t^i\}_{i \in [K]}$, we can play in determinist way, namely, $x_t = \EE(x_t^{a_t}) = \sum_{i = 1}^{T} p_t(i)x_t^i$  and thus, obtain a regret bound of $\EE(\Rcal_T)$ in \eqref{eq:prop}.
\end{corollary}

\begin{corollary}
If we choose the expert advice algorithm $\Acal$ such that $\Rcal_T^{\Acal}$ is comparable to the best of $\{\Rcal_T^{\OCO_i}\}$, then using $\Acal$ in our framework results in achieving  a regret bound that is comparable with the best optimizers in $\Mcal$.  
\end{corollary}

In order to compare these regret bounds, we need to introduce an important lemma. Thus, Lemma~\ref{lemma:lowerbound} will help us compare $\Rcal_T^{\Acal}$ and $\Rcal_T^{\OCO_i}$ appeared in proposition \ref{prop:framework}. 

\begin{lemma}[Main Lemma]\label{lemma:lowerbound}
Let $\Acal$ be an arbitrary OCO algorithm that performs on $L$-Lipschitz cost functions, w.r.t. some norm $\|\cdot\|_.$, and domains with diameter $D$, w.r.t. the same norm. Then, the regret bound for this algorithm, \ie, $\Bcal_T^\Acal$, is lower bounded by $\Omega(LD\sqrt{T})$.
\end{lemma}

It should be emphasized that in Framework \ref{alg:framework}, the availability of feedback is in our control by choosing $S$ as  $\{\ell_t(i)\}_{i \in [K]}$ or $\ell_t(a_t)$. In fact choice of $S$  is based on the full or partial feedback property of $\Acal$. 
Note that although having limited feedback might result in an increase of regret, but it also causes a reduction in computational complexity of the proposed algorithm. In \S~\ref{sec:MGD} and \S~\ref{sec:FMGD}, we will elaborate more on this trade-off. In the following, we exploit two choices of expert advice algorithms, namely, Squint and GBPA, which result in proposing \masteralg (MGD) and \fastmasteralg (FMGD), respectively.

\subsection{\masteralg}\label{sec:MGD}
Consider Framework~\ref{alg:framework} with Squint as the expert advice algorithm. We call this algorithm \masteralg (MGD) that is described in Algorithm~\ref{alg:squint-master}.
  
\begin{algorithm}[h]
\caption {\masteralg (MGD)}
\begin{algorithmic}\label{alg:squint-master}
    \STATE {\textbf{Input:} Learning rate $\eta > 0$, family of optimizers $\mathcal{M} = \{\OCO_i\}_{i\in [K] }$ with update rules $\{\Psi_i\}_{i \in [K]}$}
    \vspace{-.3cm}
    \STATE {\textbf{Initialization:} Let $R_0, V_0 \in \RR^K , x_0 \in \RR^d$ be all-zero vectors, $p_1 \in \simplex(K)$ be uniform distribution}
    \vspace{-.35cm}
    \FOR{$t = 1, \ldots, T$}
    \FOR{$a = 1, \ldots, k$}
     \STATE {Run the $a$-th optimizer algorithm and attain $x_t^a = \Psi_a(x_{t-1}^a, \grad_{1:t-1})$}
     \ENDFOR
      \STATE {Play $x_t = \sum_{a=1}^k p_t(a)x_t^a$ and observe cost function $f_t(x)$}  
      \STATE {Pass surrogate cost function $\widehat{f}_t(x) = \inner{\grad_t}{x}$  to all optimizers}
     \STATE {Set loss for the $a$-th expert  as: $\ell_t(a) = \frac{\widehat{f}_t(x_t^a)}{2F}+\frac{1}{2}$}
     \STATE{Let  $\forall i : r_t(i)=  \inner{p_t}{\ell_t} - \ell_t(i)$ then update $ R_t = R_{t-1}+r_t$ , $\forall i : V_t(i) = V_{t-1}(i) + r_t(i)^2$}
          \STATE {Compute $p_{t+1} \in \Delta(K)$ such that $p_{t+1}(i) \propto p_1(i) \exp(-\eta R_{t}(i)) + \eta^2 V_{t}(i))$}
    \ENDFOR
\end{algorithmic}
\end{algorithm}
 
Based on Proposition \ref{prop:framework} and Lemma \ref{lemma:lowerbound}, we can provide a regret bound for the MGD algorithm, as stated in Theorem~\ref{theorem:master}. The detailed proof is provided in \Apn~\ref{appendix:analysis}.

\begin{theorem}\label{theorem:master}
 Consider  MGD algorithm with a set of \ogdlike optimizers $\mathcal{M}$. Suppose $\OCO_i$ is an arbitrary optimizer  in $\Mcal$  under assumptions stated in \S~\ref{sec:assumptions}, MGD ensures
\begin{equation*}
	\Rcal_T \leq   4F\sqrt{V_T(i)\ln K} + \Rcal_T^{\OCO_i}  = \mathcal{O}\left( \sqrt{\ln K} \Bcal_T^{\OCO_i} \right),
\end{equation*}
where $F$ is the tight upper bound for all cost functions, $\Bcal_T^{\OCO_i}$ is also the tight regret upper bound  for $\OCO_i$ algorithm and $V_T(i) = \sum_{t =1}^{T}  (\inner{p_t}{\ell_t} - \ell_t(i))^2$.
\end{theorem}
\begin{remark}
If we use Hedge as expert advice algorithm in Framework \ref{alg:framework} nstead of Squint,  it achieve the same regret bound as Theorem \ref{theorem:master}.	
\end{remark}
\begin{remark}
The value of $V_T(i)$ can be much smaller than $T$, so the regret of MGD can be bounded by the best regret among all optimizers. 
\end{remark}
\begin{corollary}
 Theorem~\ref{theorem:master} shows that the \masteralg framework gives a comparable regret bound with the best algorithms of $\Mcal$ in hindsight.
\end{corollary}

\subsection{\fastmasteralg}\label{sec:FMGD}
Although MGD only needs one oracle access to cost functions $\{f_t\}_{t\in [T]}$, it needs to apply update rules of all $K$ optimizers simultaneously in each iteration. So its computational cost is higher than a \ogdlike algorithm. However, if the closed form of the update rule $\Upsilon$ can be computed efficiently same as computing iterative update rule $\Psi$, then we can provide an algorithm that can effectively reduce MGD time complexity up to factor $\frac{1}{K}$. 

We will show that the proposed algorithm, named \fastmasteralg\hspace{-2pt}, achieves almost same regret bound as MGD. This algorithm is obtained from Framework~\ref{alg:framework} in partial feedback setting that uses GBPA as its expert advice algorithm.  ‌GBPA uses Tsallis entropy $S_\alpha(x)$  and its Fenchel conjugate  which is introduced in \Apn~\ref{appendix:expert}. According to Corollary \ref{corollary:exp3},  if $\alpha \rightarrow 1$ then EXP3 is also covered.  The details of FMGD is described in Algorithm~\ref{alg:fastmaster}. We provide a regret bound for FMGD, as stated in Theorem~\ref{theorem:Fast_master}. The detailed proof is provided in \Apn~\ref{appendix:analysis}. 
 
 \begin{algorithm}[h]
\caption {\fastmasteralg (FMGD)}
\begin{algorithmic}\label{alg:fastmaster}
    \STATE {\textbf{Input:} Learning rate $\eta > 0$, family of optimizers $\mathcal{M} = \{\OCO_i\}_{i\in [K] }$ with closed form update rules $\{\Upsilon_i\}_{i \in [K]}$ }
    \STATE {\textbf{Initialization:} Let $\widehat{L}_0 \in \RR^K, x_0 \in \RR^d$ be all-zero vectors and $p_1 \in \simplex(K)$ be the uniform distribution over the family of optimizers $\Mcal$}
    \FOR{$t = 1, \ldots, T$}
     \STATE {Choose $a_t \sim p_t$ as an action}
     \STATE {Run the $a_t$-th expert algorithm and attain $x_t = \Upsilon_{a_t}(x_{0},\grad_{1:t-1})$}
     \STATE {Observe cost function $f_t$}
     \STATE {Set loss for the $a_t$-th expert as: $\ell_t(a_t) = \frac{\widehat{f}_t(x_t)}{2F}+\frac{1}{2} = \frac{\inner{\grad_t}{x_t}}{2F}+\frac{1}{2}$}
     \STATE{Update $\widehat{L}_t = \widehat{L}_{t-1}+\widehat{\ell}_t$ where $\forall a \in [K]  : \widehat{\ell}_t(a) = \frac{\ell_t(a)}{p_t(a)} \textbf{1} \{ a = a_t \} $}
     \STATE {Compute $p_{t+1} \in \Delta(K)$ such that $p_{t+1} \propto \grad S_\alpha^*\left( -\frac{ \widehat{L}_t }{\eta} \right)$}
    \ENDFOR 
\end{algorithmic}
\end{algorithm}

\begin{theorem}\label{theorem:Fast_master}
Consider FMGD algorithm with optimizer set $\mathcal{M}$ that consists of \ogdlike optimizers.  Then for all optimizers $\OCO_i \in \Mcal$, under assumptions  stated in \S~\ref{sec:assumptions}, FMGD ensures
\begin{equation*}
\begin{aligned}
	&\alpha = 1/2\  (GBPA):   \EE(\Rcal_T) \leq  8F\sqrt{T  K} +  \Bcal_T^{\OCO_i}  =\mathcal{O}\left( \sqrt{K} \Bcal_T^{\OCO_i}\right),\\
		&\alpha \rightarrow 1 \ (EXP3 )  : \EE(\Rcal_T) \leq  2F\sqrt{2T K \ln K} + \Bcal_T^{\OCO_i} = \mathcal{O}\left( \sqrt{K\ln K} \Bcal_T^{\OCO_i}\right),
	\end{aligned}
\end{equation*}
where $F$ is a tight upper bound for all surrogate cost functions and  $\Bcal_T^{\OCO_i}$ is the tight upper bound regret for $\OCO_i$ algorithm.
\end{theorem}
\begin{corollary}\label{corollary:fast}
In regret bound term, FMGD attains same regret bound as  MGD (differs by at most $\sqrt{K / \ln K}$ multiplicative factor). In computational terms, if for each members of $\Mcal$, the closed form update rule $\Upsilon$ can be computed with the same complexity as $\Psi$, then in the worst case FMGD achieves the same complexity as the worst complexity of algorithms in $\Mcal$. Hence, its computational complexity is improved by a multiplicative factor of $\frac{1}{K}$.	
\end{corollary}
 
\subsection{Learning The Best Regularizer}\label{sec:bestReg}
Consider the problem described in \S~\ref{sec:ProbState}, where we have $K$ lazy-OMD algorithms (described in Definition~\ref{def:MD}) that are determined by $K$ different regularizer functions. Now, in order to compete with the best regularizer, we can take advantage of MGD algorithm with its optimizers set $\Mcal$ consisting of lazy-OMD algorithms. According to Proposition~\ref{prop:ComputeUpdateRuleMD}, closed form of update rules for lazy-OMD algorithms, can be computed efficiently by  keeping track of $S_t = \sum_{i=1}^t \grad_i$ in each iteration. Consequently, based on what is stated in Corollary \ref{corollary:fast},  using FMGD leads to learning the best regularizer with low computational cost.

Now in Theorem~\ref{theorem:best-regularizer} we express our results on learning the best regularizer among a family of regularizers.

\begin{theorem}\label{theorem:best-regularizer}
Let $\Phi$ be a set of $K$ regularizers in which the $i$-th member $\varphi_i \colon \Dcal \to \RR$ is $\rho_i$-strongly convex w.r.t. a norm $\|\cdot \|_i$. Let $D_i = \sup_{x\in \Xcal} B_{\varphi_i}(x,x_0)$ where $B_{\varphi_i}(x,x_0) =   \varphi_i(x) - \inner{\grad \varphi_i(x_0)}{x-x_0} - \varphi_i(x_0)$. Let cost functions $\{f_t\}_{t\in [T]}$ be convex and $L_i$-Lipschitz w.r.t. $\|\cdot\|_{i}$ and upper bounded by $F$. Then for any $i \in [K]$, our proposed algorithms MGD and FMGD ensure
\begin{equation*}
\begin{aligned}
	\text{MGD:} ~~~  &\Rcal_T \leq 4F\sqrt{T \ln K} + L_i\sqrt{2D_iT/\rho_i} \leq  (4\sqrt{\ln K} + 1) L_i\sqrt{2D_iT/\rho_i},\\
	\text{FMGD:}~~~ & \EE(\Rcal_T) \leq 8F\sqrt{T K } + L_i\sqrt{2D_i T/\rho_i} \leq  (8\sqrt{K} + 1) L_i\sqrt{2D_iT/\rho_i}.
\end{aligned}
\end{equation*}
\end{theorem}

\begin{remark}
The computational complexity of MGD is at most $K$ times more costly than that of a lazy-OMD and the complexity of FMGD is the same as a lazy-OMD. Both FMGD and MGD algorithms only need one oracle access to cost function per iteration.
\end{remark}

		\section{Experimental Results}
\label{sec:ExpResults}
In this section, we demonstrate the practical utility of our proposed Framework \ref{alg:framework}. Toward this end, we present an experiment that fits a linear regression model on synthetic data with square loss. In this experiment, we compare MGD and FMGD with a family of lazy-OMD algorithms in terms of average regret. Finally, we compare  the execution time of MGD and FMGD. To support our results, in \Apn~\ref{appendix:domain}, a comparision between  negative entropy and quadratic regularizer for $\BB_2$ and $\simplex(d)$ to find the best regularizer has been performed.

\subsection{Learning the Best Regularizer for Online Linear Regression}
\label{sec:ExperimentDetails}
In the first set of experiments, we preform an online linear regression model \cite{auer2002adaptive} on a synthetic dataset which has been generated in the following way. Let the feature vector $x_t \in \mathbb{R}^{20}$ be sampled from a truncated multivariate normal distribution. Additionally,  a weight $w$ is sampled uniformly at random from $\BB_2$. The value associated with the feature vector $x_t$ is set by $y_t = \inner{w}{x_t} +\epsilon$ where $\epsilon \sim \mathcal{N}(0,1)$. The model is trained and evaluated against square loss. As mentioned in \S~\ref{sec:bestReg}, we consider that the experts set of MGD and FMGD consists of an OMD family with different choices of regularizers. Also, it should be mentioned that we use Hedge algorithm for expert tracking in MGD and Exp3 algorithm in FMGD.
We have trained the above regression problem using our proposed  framework, described in \S~\ref{proposed}, for the following two cases.

$\boldsymbol{\BB_2}$ \textbf{Domain:}
 In the first case, we trained the model over the probability simplex. The family of experts $\Mcal$ contains 8 OMD algorithms using Hypentropy \cite{HypEnt} regularizer where the parameter  $\beta$ is chosen from  $\{2^n: -5 \leq n \leq 2 , n \in \mathbb{Z}\}$. Moreover, the experts family contains an OMD with quadratic regularizer and  another OMD with negative entropy regularizer.

\textbf{Simplex Domain:}
In the second case, we trained the model over $\BB_2$. Here, we consider a family of experts that contain 8 OMD algorithms using Hypentropy regularizer with parameter $\beta$  chosen from $\{2^n: -4 \leq n \leq 3 , n \in \mathbb{Z}\}$, and an OMD with quadratic regularizer. 

\begin{figure}
	\centering
	\begin{subfigure}{.5\textwidth}
		\centering
		\includegraphics[width=.94\linewidth, height= 110pt]{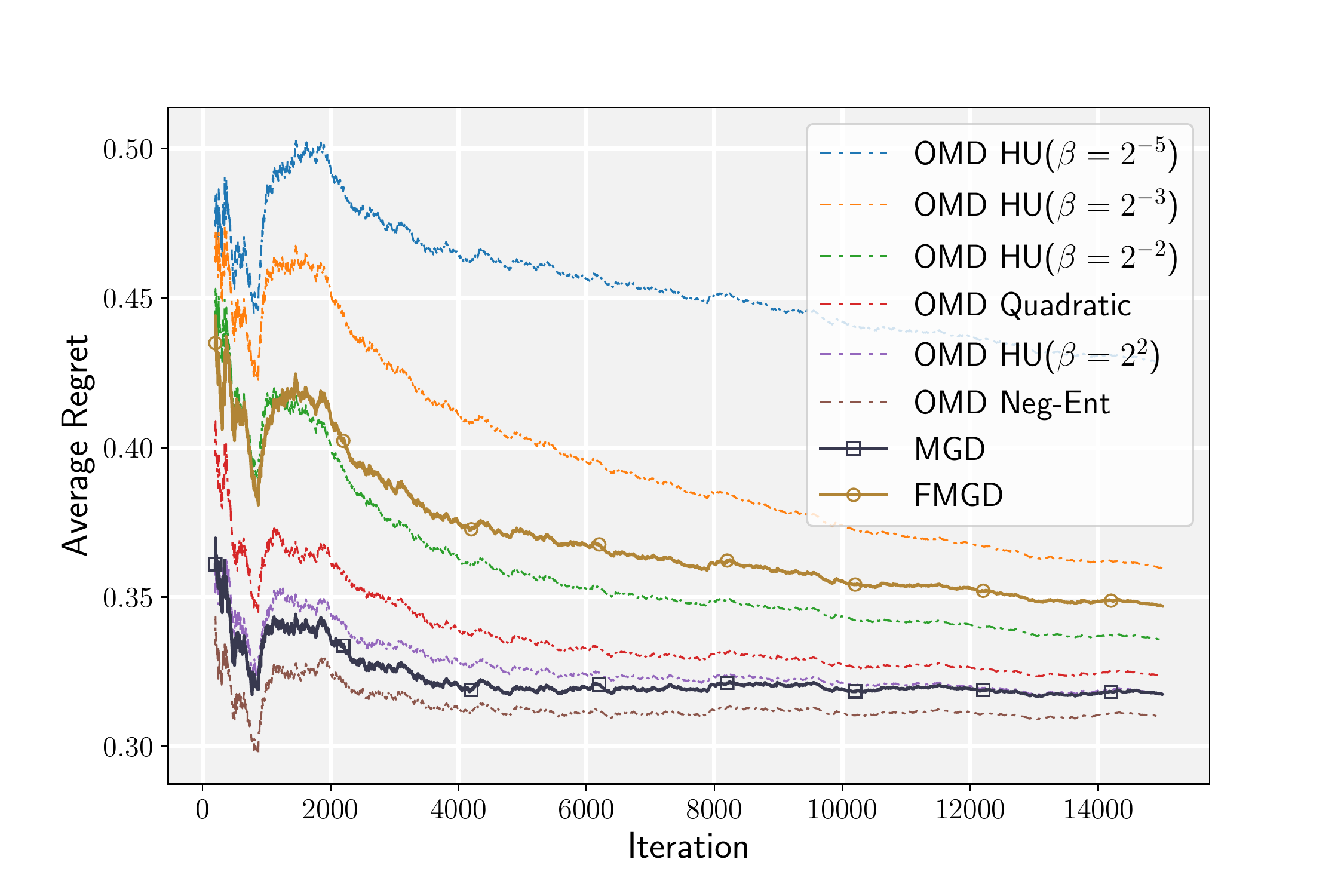}
		\caption{}
		\label{fig:RegretSim}
	\end{subfigure}%
	\begin{subfigure}{.5\textwidth}
		\centering
		\includegraphics[width=.94\linewidth, height= 110pt]{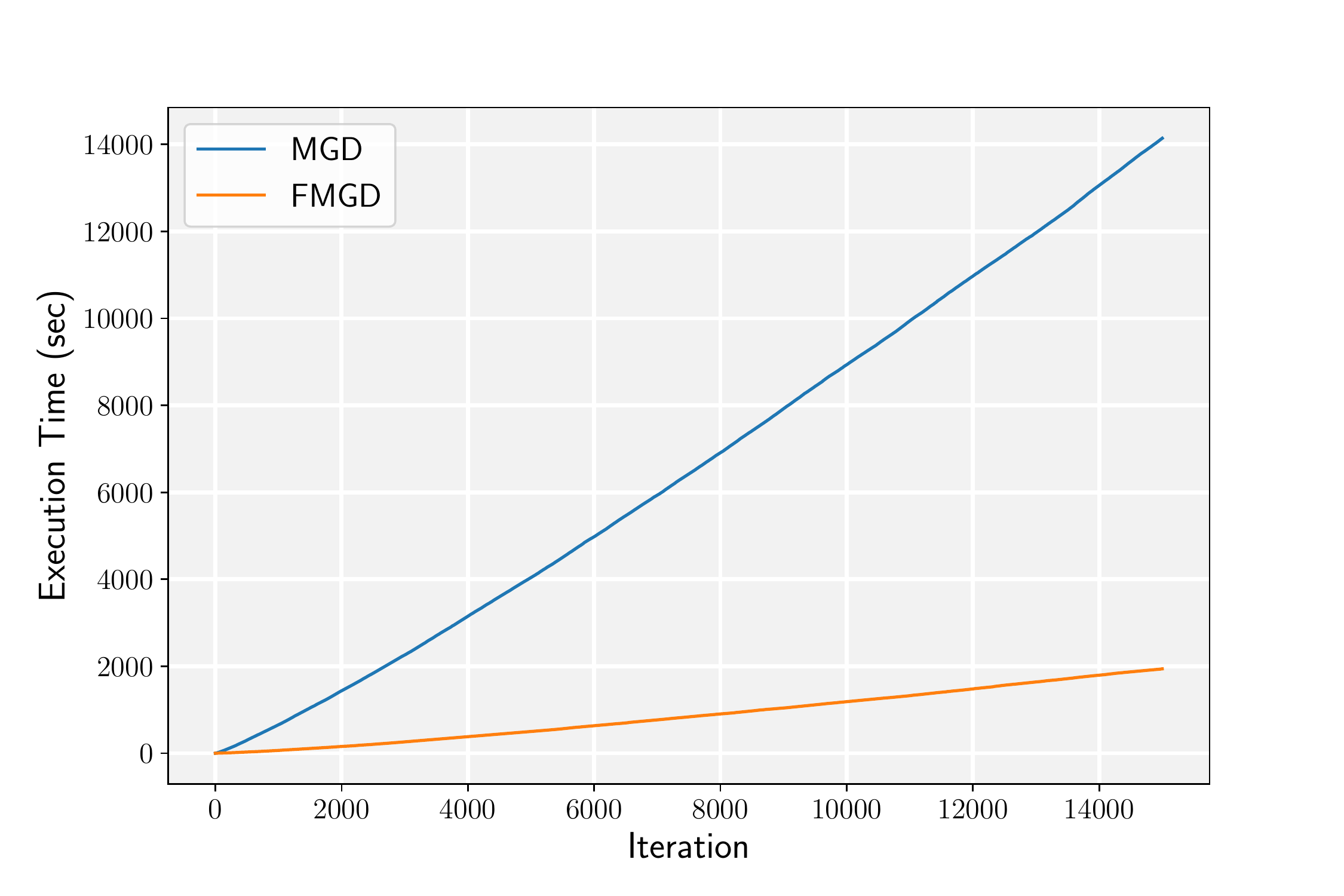}
		\caption{}
		\label{fig:ExecTimeSimp}
	\end{subfigure}%

	\begin{subfigure}{.5\textwidth}
	\centering
	\includegraphics[width=.94\linewidth, height= 110pt]{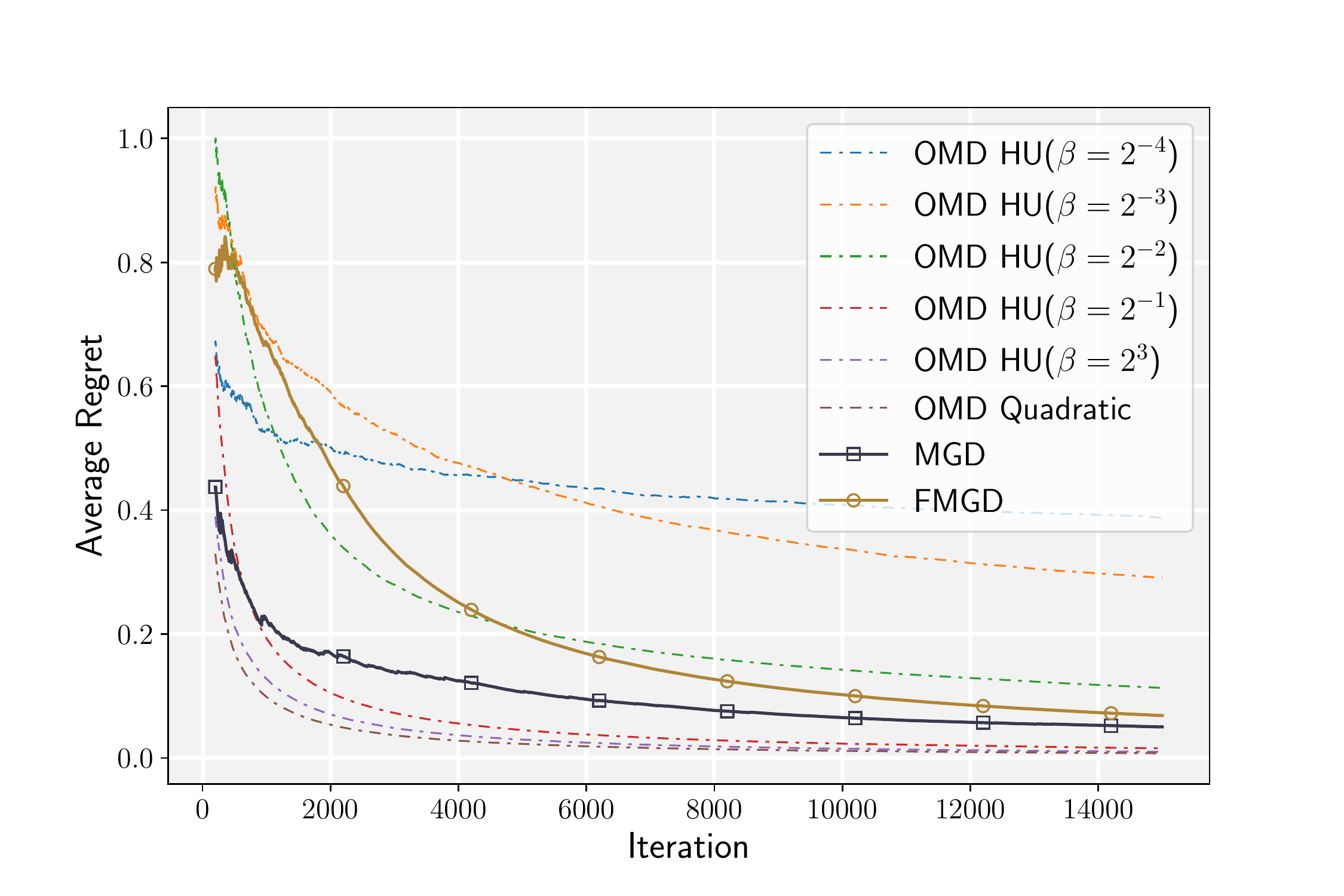}
	\caption{}
	\label{fig:RegretL2b}
	\end{subfigure}%
	\begin{subfigure}{.5\textwidth}
	\centering
	\includegraphics[width=.94\linewidth, height= 110pt]{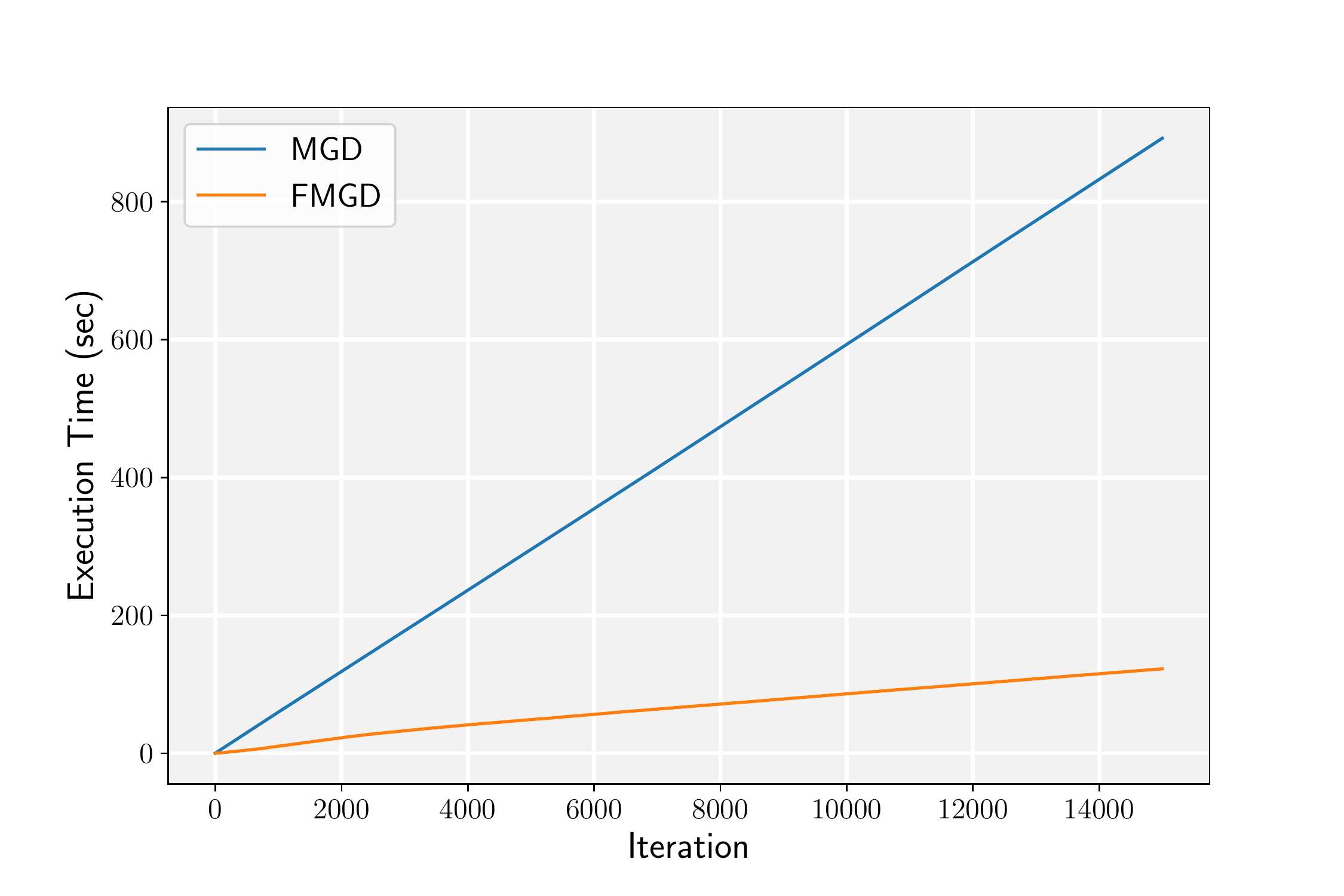}
	\caption{}
	\label{fig:ExecTimeL2b}
	\end{subfigure}%

	\caption{The top row and bottom row demonstrate experimental results for the proposed framework on simplex and $\BB_2$, respectively. The details of experiments are described in \S~\ref{sec:ExperimentDetails}.}
	\label{fig:Result}
\vspace{-20pt}
\end{figure}

\textbf{Results:}
The results of experiments mentioned above are demonstrated in Figure \ref{fig:Result}. We have computed the average regret and have used it as a measure to compare the performance of OCO algorithms. The top row and the bottom row of Figure \ref{fig:Result}  depict the results of optimization over simplex domain and $\BB_2$ domain, respectively. Figures \ref{fig:RegretSim} and \ref{fig:RegretL2b} illustrate the change in average regret with respect to time. The results closely track those predicted by the theory, as stated in Theorem~\ref{theorem:best-regularizer}. Besides, it can be seen that OMD with a negative entropy regularizer in the simplex domain case, and OMD with a quadratic regularizer in the $\BB_2$ domain case outperform other regularizers. It can also be noted that in both cases MGD performs  closely to the best regularizer and  FMGD performs reasonably well. Figures \ref{fig:ExecTimeSimp} and \ref{fig:ExecTimeL2b} investigate the running time of MGD and FMGD. As expected, the time ratio between MGD and FMGD is a constant, approximately equal to the size of experts set.

		\section{Discussion and Future Work}
In this paper, we have investigated the problem of finding the best algorithm among a class of OCO algorithms. To this end, we introduced a novel framework for OCO, based on the idea of employing \emph{expert advice} and \emph{bandit} as a master algorithm. 
As a special case, one can choose the family of optimizers based on the step size. In this case, the MetaGrad algorithm \cite{van2016metagrad} can be recovered as a special case of our framework. 
Furthermore, we can choose the family of optimizers based on parameters about which we usually have no information such as Lipschitz constant, domain diameter, strong convexity coefficient, \etc~
In this work, the family of OCO algorithms are considered to be finite. An interesting direction for future work would be to investigate the problem setup for a family of infinite algorithms. 
Moreover, we showed that partial and full feedback approaches maintain a trade-off between complexity and regret bound. As another potential direction for future work, one can consider the case of using feedback from more than one experts.
From the enviornment’s point of view, we have studied the static regret. However, it should be emphasized that the dynamic regret \cite{hall2013dynamical,jadbabaie2015online,mokhtari2016online, yang2016tracking,zhang2017improved} can be analyzed in the same fashion. 
Finally, to get the results of our experiments, stated in \S~\ref{sec:ExpResults}, we have used EXP3 in partial feedback setting. However, in practice we believe that employing algorithms more suitable for stochastic environment \cite{bubeck12regret} like \emph{Thompson sampling} \cite{russo2018tutorial} may lead to even better results.

		
		
		\bibliography{neurips_2019.bib}

\begin{thebibliography}{24}
\providecommand{\natexlab}[1]{#1}
\providecommand{\url}[1]{\texttt{#1}}
\expandafter\ifx\csname urlstyle\endcsname\relax
  \providecommand{\doi}[1]{doi: #1}\else
  \providecommand{\doi}{doi: \begingroup \urlstyle{rm}\Url}\fi

\bibitem[Abernethy et~al.(2015)Abernethy, Lee, and
  Tewari]{abernethy2015fighting}
Jacob~D Abernethy, Chansoo Lee, and Ambuj Tewari.
\newblock Fighting bandits with a new kind of smoothness.
\newblock In \emph{Advances in Neural Information Processing Systems}, pages
  2197--2205, 2015.

\bibitem[Auer et~al.(2002{\natexlab{a}})Auer, Cesa-Bianchi, Freund, and
  Schapire]{auer2002nonstochastic}
Peter Auer, Nicolo Cesa-Bianchi, Yoav Freund, and Robert~E Schapire.
\newblock The nonstochastic multiarmed bandit problem.
\newblock \emph{SIAM journal on computing}, 32\penalty0 (1):\penalty0 48--77,
  2002{\natexlab{a}}.

\bibitem[Auer et~al.(2002{\natexlab{b}})Auer, Cesa-Bianchi, and
  Gentile]{auer2002adaptive}
Peter Auer, Nicolo Cesa-Bianchi, and Claudio Gentile.
\newblock {Adaptive and self-confident on-line learning algorithms}.
\newblock \emph{Journal of Computer and System Sciences}, 64:\penalty0 48--75,
  2002{\natexlab{b}}.

\bibitem[Bubeck et~al.(2012)Bubeck, Cesa-Bianchi, et~al.]{bubeck12regret}
Sebastien Bubeck, Nicolo Cesa-Bianchi, et~al.
\newblock Regret analysis of stochastic and nonstochastic multi-armed bandit
  problems.
\newblock \emph{Foundations and Trends{\textregistered} in Machine Learning},
  5\penalty0 (1):\penalty0 1--122, 2012.

\bibitem[Cesa-Bianchi and Lugosi(2006)]{bianchi-2006-prediction}
Nicolo Cesa-Bianchi and Gabor Lugosi.
\newblock \emph{Prediction, Learning, and Games}.
\newblock Cambridge University Press, New York, NY, USA, 2006.

\bibitem[Cutkosky and Boahen(2017)]{cutkosky2017online}
Ashok Cutkosky and Kwabena Boahen.
\newblock Online learning without prior information.
\newblock \emph{arXiv preprint arXiv:1703.02629}, 2017.

\bibitem[Duchi et~al.(2011)Duchi, Hazan, and Singer]{JMLR:Adaptive}
John Duchi, Elad Hazan, and Yoram Singer.
\newblock Adaptive subgradient methods for online learning and stochastic
  optimization.
\newblock \emph{Journal of Machine Learning Research}, 12\penalty0
  (Jul):\penalty0 2121--2159, 2011.

\bibitem[Freund and Schapire(1997)]{Freund:1997}
Yoav Freund and Robert~E Schapire.
\newblock A decision-theoretic generalization of on-line learning and an
  application to boosting.
\newblock \emph{Journal of computer and system sciences}, 55\penalty0
  (1):\penalty0 119--139, 1997.

\bibitem[Ghai et~al.(2019)Ghai, Hazan, and Singer]{HypEnt}
Udaya Ghai, Elad Hazan, and Yoram Singer.
\newblock Exponentiated gradient meets gradient descent.
\newblock \emph{arXiv preprint arXiv:1902.01903}, 2019.

\bibitem[Hall and Willett(2013)]{hall2013dynamical}
Eric~C Hall and Rebecca~M Willett.
\newblock Dynamical models and tracking regret in online convex programming.
\newblock \emph{arXiv preprint arXiv:1301.1254}, 2013.

\bibitem[Hazan et~al.(2007)Hazan, Agarwal, and Kale]{ML:Hazan:2007}
Elad Hazan, Amit Agarwal, and Satyen Kale.
\newblock Logarithmic regret algorithms for online convex optimization.
\newblock \emph{Machine Learning}, 69\penalty0 (2-3):\penalty0 169--192, 2007.

\bibitem[Hazan et~al.(2016)]{Intro:Online:Convex}
Elad Hazan et~al.
\newblock Introduction to online convex optimization.
\newblock \emph{Foundations and Trends{\textregistered} in Optimization},
  2\penalty0 (3-4):\penalty0 157--325, 2016.

\bibitem[Jadbabaie et~al.(2015)Jadbabaie, Rakhlin, Shahrampour, and
  Sridharan]{jadbabaie2015online}
Ali Jadbabaie, Alexander Rakhlin, Shahin Shahrampour, and Karthik Sridharan.
\newblock Online optimization: Competing with dynamic comparators.
\newblock In \emph{Artificial Intelligence and Statistics}, pages 398--406,
  2015.

\bibitem[Koolen and Van~Erven(2015)]{squint}
Wouter~M Koolen and Tim Van~Erven.
\newblock Second-order quantile methods for experts and combinatorial games.
\newblock In \emph{Conference on Learning Theory}, pages 1155--1175, 2015.

\bibitem[Littlestone and Warmuth(1994)]{LittlestoneWarmuth1994}
Nick Littlestone and Manfred~K Warmuth.
\newblock The weighted majority algorithm.
\newblock \emph{Information and computation}, 108\penalty0 (2):\penalty0
  212--261, 1994.

\bibitem[McMahan and Streeter(2010)]{McMahanStreeter2010}
H~Brendan McMahan and Matthew Streeter.
\newblock Adaptive bound optimization for online convex optimization.
\newblock \emph{arXiv preprint arXiv:1002.4908}, 2010.

\bibitem[Mokhtari et~al.(2016)Mokhtari, Shahrampour, Jadbabaie, and
  Ribeiro]{mokhtari2016online}
Aryan Mokhtari, Shahin Shahrampour, Ali Jadbabaie, and Alejandro Ribeiro.
\newblock Online optimization in dynamic environments: Improved regret rates
  for strongly convex problems.
\newblock In \emph{2016 IEEE 55th Conference on Decision and Control (CDC)},
  pages 7195--7201. IEEE, 2016.

\bibitem[Russo et~al.(2018)Russo, Van~Roy, Kazerouni, Osband, Wen,
  et~al.]{russo2018tutorial}
Daniel~J Russo, Benjamin Van~Roy, Abbas Kazerouni, Ian Osband, Zheng Wen,
  et~al.
\newblock A tutorial on thompson sampling.
\newblock \emph{Foundations and Trends{\textregistered} in Machine Learning},
  11\penalty0 (1):\penalty0 1--96, 2018.

\bibitem[Shalev-Shwartz et~al.(2012)]{Online:suvery}
Shai Shalev-Shwartz et~al.
\newblock Online learning and online convex optimization.
\newblock \emph{Foundations and Trends{\textregistered} in Machine Learning},
  4\penalty0 (2):\penalty0 107--194, 2012.

\bibitem[van Erven and Koolen(2016)]{van2016metagrad}
Tim van Erven and Wouter~M Koolen.
\newblock Metagrad: Multiple learning rates in online learning.
\newblock In \emph{Advances in Neural Information Processing Systems}, pages
  3666--3674, 2016.

\bibitem[Vovk(1998)]{Vovk1998}
Vladimir Vovk.
\newblock A game of prediction with expert advice.
\newblock \emph{Journal of Computer and System Sciences}, 56\penalty0
  (2):\penalty0 153--173, 1998.

\bibitem[Vovk(1990)]{vas-90}
Volodimir~G Vovk.
\newblock Aggregating strategies.
\newblock \emph{Proc. of Computational Learning Theory, 1990}, 1990.

\bibitem[Yang et~al.(2016)Yang, Zhang, Jin, and Yi]{yang2016tracking}
Tianbao Yang, Lijun Zhang, Rong Jin, and Jinfeng Yi.
\newblock Tracking slowly moving clairvoyant: optimal dynamic regret of online
  learning with true and noisy gradient.
\newblock In \emph{Proceedings of the 33rd International Conference on
  International Conference on Machine Learning-Volume 48}, pages 449--457.
  JMLR. org, 2016.

\bibitem[Zhang et~al.(2017)Zhang, Yang, Yi, Rong, and Zhou]{zhang2017improved}
Lijun Zhang, Tianbao Yang, Jinfeng Yi, Jing Rong, and Zhi-Hua Zhou.
\newblock Improved dynamic regret for non-degenerate functions.
\newblock In \emph{Advances in Neural Information Processing Systems}, pages
  732--741, 2017.

\end{thebibliography}
		\bibliographystyle{plain}
	\newpage
		\appendix	
		\section{Background}
In this section definition  of Expert advice algorithm provided. After that Bregman Divergence definition,  which is used in OMD algorithms, is  provided. Then OMD algorithm and regret bound of it, is mentioned. 
\subsection{Expert Advice}\label{appendix:expert}
On expert advice we have discussed but the framework and detailed algorithm of them are not provided. In this section some of algorithms in expert advice problem that we have used are introduced in detailed.
\subsection{Framework}
Expert advice framework:
\begin{algorithm}[H]
\caption {Expert Advice}
\begin{algorithmic}
    \STATE {\textbf{Input:} Learning rate $\eta > 0$}
    \STATE {\textbf{Initialization:} Let $p_1$ be the distribution, according to the prior knowledge about experts}
    \FOR{$t = 1, \ldots, T$}
    \STATE {Get all experts predictions and play $a_t$ based on $p_t$ and predictions}
     \STATE {Observe losses of all experts as the vector $\ell_t$ }
     \STATE {Update $p_{t+1} \in \Delta(K)$ based on losses we have observed so far}
    \ENDFOR
\end{algorithmic}
\end{algorithm}
All of the below algorithms follow the above framework.
\subsection{Squint}
Squint algorithm is stated as bellow.
\begin{algorithm}[H]
\caption {Squint Algorithm}
\begin{algorithmic}
    \STATE {\textbf{Input:} learning rate $\eta > 0$}
    \STATE {\textbf{Initialization:} let $R_0, V_0$ be two all-zero vectors}
    \FOR{t = 1, \ldots, T}
     \STATE {compute $p_t \in \Delta(K)$ such that $p_t(a) \propto p_1(i) exp(-\eta R_{t-1}(a)) + \eta^2 V_{t-1})$}
     \STATE {play $a_t \sim p_t$ and observe its loss vector $\ell_t$}
     \STATE{update $R_t = R_{t-1}+\ell_t$ and $\forall i : V_t(i) = V_{t-1}(i) + \ell_t(i)^2$}
    \ENDFOR
\end{algorithmic}
\label{alg:squint}
\end{algorithm}
In fact in above algorithm, $R_t(a)$ denotes the expected regret w.r.t. $a$-th expert. 
\subsection{GBPA}
GBPA algorithm is defined as bellow.
\begin{algorithm}[H]
\caption {GBPA}
\begin{algorithmic}
    \STATE {\textbf{Input:} learning rate $\eta > 0$}
    \STATE {\textbf{Initialization:} let $\widehat{L}_0$ be the all-zero vector}
    \FOR{t = 1, \ldots, T}
     \STATE {compute $p_t \in \Delta(K)$ such that $p_t \propto \grad S_{\alpha}^*( - 
     \frac{\widehat{L}_{t-1}}{\eta})$}
     \STATE {play $a_t \sim p_t$ and observe its loss $\ell_t(a_t)$}
     \STATE{update $\widehat{L}_t = \widehat{L}_{t-1}+\widehat{\ell}_t$ where $\widehat{\ell}_t(a) = \frac{\ell_t(a)}{p_t(a)} \textbf{1} \{ a = a_t \} , \forall a \in [K] $}
    \ENDFOR 
    \end{algorithmic}
    \end{algorithm}
    Above algorithm uses Tsallis entropy which is defined as below.
    \begin{equation}\label{eq:tsallis}
    S_{\alpha}(L) = \frac{1}{1-\alpha}(1 - \sum\limits_{i=1}^{K} L(i)^{\alpha})
    \end{equation}
    EXP3 algorithm is GBPA where in \eqref{eq:tsallis} $\alpha \rightarrow 1$. Now we want to compute its update rule of probabilities for EXP3. By  using L'Hôpital's rule, we have
   \[
   \begin{aligned}
   	   \lim\limits_{\alpha \rightarrow 1} S_{\alpha}(L) = & \lim_{\alpha \rightarrow 1} \frac{(1- \sum_{i=1}^{K} L(i)^{\alpha})'}{(1-\alpha)'}\\
   	   = &  \lim_{\alpha \rightarrow 1} \frac{ \sum_{i=1}^{K} -\ln L(i) \cdot p_i^{\alpha}}{-1}\\
   	   = & \sum\limits_{i=1}^{K} \ln L(i) \cdot L(i) = H(L)
   	      \end{aligned}
   \]
where $H(L)$ is negative entropy function. We know that $H^*(L) = \sup\limits_{x \in \RR^K}(\inner{L}{x} - H(x))$ so:
\[
H^*(\frac{L}{\eta}) =  \frac{1}{\eta} \ln(\sum\limits_{i=1}^{K} \exp(\eta L(i)))
\]
 So $p_t(a)$ is the $a$-th element of $\grad H^*(-\frac{\widehat{L}_{t-1}}{\eta})$   which is : $\frac{\exp(-\eta\widehat{L}_{t-1}(a))}{\sum_{i=1}^{K} \exp(-\eta\widehat{L}_{t-1}(i))}$. So EXP3 is defined by following algorithm.
\begin{algorithm}[H]
\caption {EXP3}
\begin{algorithmic}
    \STATE {\textbf{Input:} learning rate $\eta > 0$}
    \STATE {\textbf{Initialization:} let $\widehat{L}_0$ be the all-zero vector}
    \FOR{t = 1, \ldots, T}
     \STATE {compute $p_t \in \Delta(K)$ such that $p_t(a) \propto exp(-\eta \widehat{L}_{t-1}(a))$}
     \STATE {play $a_t \sim p_t$ and observe its loss $\ell_t(a_t)$}
     \STATE{update $\widehat{L}_t = \widehat{L}_{t-1}+\widehat{\ell}_t$ where $\widehat{\ell}_t(a) = \frac{\ell_t(a)}{p_t(a)} \textbf{1} \{ a = a_t \} , \forall a \in [K] $}
    \ENDFOR 
    \end{algorithmic}
    \end{algorithm}

\subsection{Bregman Divergence}
Let $F\colon \Dcal\subset \RR^d\to \RR$ be a strictly convex and differentiable function. Denote by $B_F(x,y)$ the Bregman divergence associated with $F$ for points $x,y$, defined by
\begin{equation}\label{eq:bregmandef}
    B_F(x,y) \triangleq F(x) - F(y) - \inner{\grad F(y)}{x-y}.
\end{equation}

We also define the projection of a point $x\in \Dcal$ onto a set $\Xcal\subset \Dcal$ with respect to $B_F$ as
\[
    \proj{\Xcal}{F}(x) \triangleq \argmin_{z\in\Xcal} B_F(z, x).
\]

Here we give some useful property of the Bregman divergence.

\begin{lemma}
    Let $F\colon \simplex(d) \to \RR_+$ be the \emph{negative entropy function}, defined as 
        $F(x) = \sum_{i=1}^d x^i\log x^i$.
    Then, we have
    \begin{equation}\label{eq:bregkl}
        B_F(x,y) = \kl{x}{y}.
    \end{equation}
    Moreover, if one extends the domain of $F$ to $\RR_+^d$, then, defining the extended KL divergence as 
    \[
        \kl{x}{y} = \sum_{i=1}^d x^i\log\frac{x^i}{y^i} - (x^i - y^i),
    \]
    the equality \eqref{eq:bregkl} holds.
\end{lemma}

\subsection{Mirror Descent}
The Online Mirror Descent (OMD) algorithm is defined as follows. Let $\Dcal$ be a domain containing $\Xcal$, and $\varphi\colon \Dcal \to \RR$ be a \emph{mirror map}. Let $x_1 = \argmin_{x\in \Xcal} \varphi(x)$. For $t\geq 1$, set $y_{t+1}\in \Dcal$ such that
\[
    \grad \varphi(y_{t+1}) = \grad \varphi(x_t) - \eta \grad_t,
\]
and set 
\[
    x_{t+1} = \proj{\Xcal}{\varphi}(y_{t+1}).
\]

\begin{theorem}\label{theorem:omd}
    Let $\varphi\colon \Dcal \to \RR$ be a mirror map which is $\rho$-strongly convex w.r.t. a norm $\|\cdot \|$. Let $D = \sup_{x\in \Xcal} B_{\varphi}(x,x_1)$, and $f$ be convex and $L$-Lipschitz w.r.t. $\|\cdot\|$. Then, OMD with $\eta = \frac{\sqrt{2\rho D}}{L}\frac{1}{\sqrt{T}}$ gives
\begin{equation} \label{eq:2}
      \Rcal_T \leq L\sqrt{\frac{2DT}{\rho}}.
\end{equation}
%
Note that if we did not have $T$, then if we set $\eta_t = \frac{\sqrt{2\rho D}}{L}\frac{1}{\sqrt{t}}$ we can achieve same regret bound.
\end{theorem}

\section{Analysis}\label{appendix:analysis}
Analysis of theorems and other materials in paper are stated in the following.
\subsection{Auxiliary Lemmas}
\begin{lemma}\label{lemma:l1}
Let $\Acal$ be an arbitrary expert advice algorithm, performs on expert set $\Mcal$.  Suppose that  loss of our experts have upper bound $L$ instead of being in interval $[0,1]$. Then running $\Acal$ on normalized version of losses $\bar{\ell}_t = \frac{\ell_t}{L}$ gives following regret.
\[
\Rcal_T  =	L \Rcal_T^{\Acal}
\]
where $\Rcal_T^{\Acal}$ is the regret for running algorithm $\Acal$ on normalized version of losses. 
    \begin{proof}
If we play $a_t$ at iteration $t$ then we can write Regret of our proposed algorithm on bounded losses, we can say: 
\[
\begin{aligned}	
\Rcal_T = & \sum\limits_{t = 1}^{T} \ell_t(a_t) - \min\limits_{a \in \Mcal}\sum\limits_{t = 1}^{T} \ell_t(a)\\
 = & L\left( \sum\limits_{t = 1}^{T} \bar{\ell}_t(a_t) + \min\limits_{a \in \Mcal}\sum\limits_{t = 1}^{T} \bar{\ell}_t(a)\right) =  L \Rcal_T^a 
\end{aligned}
\]
 \end{proof}
\end{lemma}

\begin{lemma}\label{lemma:l2}
	Let cost functions $\{f_t\}_{t \in [T]}$ on domain $\Dcal$ and tight upper bound for surrogate cost functions $F = \sup\limits_{x \in \Dcal, t \in [T]} |\inner{\grad_t}{x}|$. Suppose that all $f_t$ are $L$-Lipschitz w.r.t. norm $\|.\|$ and $\Dcal$ has upper bound $D$ w.r.t. norm $\|.\|$. Then $F \leq LD$.
	\begin{proof}
	We know that if $f_t$ is  $L$-Lipschitz w.r.t. norm $\|.\|$, then : $\|\grad f_t(x)\|^* \leq L$. So by the Cauchy-Schwarz Inequality we have:
\begin{align*} 
|\inner{\grad_t}{x}| & \leq \|\grad_t\|^*\|x\|\\
	& \leq LD
\end{align*}
So we have $F = \sup\limits_{x \in \Dcal, t \in [T]} |\inner{\grad_t}{x}| \leq LD$	
\end{proof} 
\end{lemma}

\subsection{Proof of Proposition 3.1}
\begin{proof}[Proof of Proposition 3.1]
Let $x^* = \arg \min\limits_{x \in \Dcal} \sum\limits_{t=1}^T \inner{\grad_t}{x}$ and $a^* = \arg \min\limits_{a \in [K]} \sum\limits_{t=1}^T \ell_t(a)$. Then for regret of our framework we have:
\[ 
\begin{aligned}
\Rcal_T & \leq \sum\limits_{t=1}^T f_t(x_t) -\sum\limits_{t=1}^T f_t(x^*) \\
& \stackrel{\text{(a)}}{\leq} \inner{\grad_t}{x_t-x^*}\\
& \stackrel{\text{(b)}}{=} \sum\limits_{t=1}^T \inner{\grad_t}{x_t^{a_t}} - 
\sum\limits_{t=1}^T \inner{\grad_t}{x_t^{a^*}}  +
 \sum\limits_{t=1}^T \inner{\grad_t}{x_t^{a^*}} -
 \sum\limits_{t=1}^T \inner{\grad_t}{x^*} \\
& =2F \left(\sum\limits_{t=1}^T \ell_t(a_t) - \sum\limits_{t=1}^T \ell_t(a^*)\right) +  \left(\sum\limits_{t=1}^T \widehat{f}_t(x_t^{a^*}) - \sum\limits_{t=1}^T \widehat{f}_t(x^*)\right)\\
& =2F \Rcal_T^{\Acal} + \Rcal_T^{\OCO_{a^*}} 
\end{aligned}
\]
where (a) follows by convexity of $\{f_t\}_{t \in [T]}$ and (b) follows by the fact that $x_t = x_t^{a_t}$. 
\end{proof}

\subsection{Proof of Lemma 3.5}
\begin{proof}[Proof of Lower Bound Lemma]
Consider	an instance of OCO where $\Kcal \subseteq \RR^d$ is a ball with diameter $D$ w.r.t norm mentioned norm.
\begin{equation}
\Kcal = \{x \in \RR^d | \|x\|_. \leq D\} = D \{x | \|x\| \leq 1\}
\end{equation}
Assume that $\bm{e}_i \in \RR^d $ be the vector  where all elements except $i$-th element are zero and the $i$-th element is $a_i > 0$ such that $\|\bm{e}_i \|_. = 1$. Define $V \triangleq \{L\bm{e}_1, \ldots, L\bm{e}_d, -L\bm{e}_1, \ldots, -L\bm{e}_d \}$ be the set of  $2d$ vectors  with norm $L$. Now define $2d$ functions as bellow:
$$\forall ~ v \in V : f_v(x) = \inner{v}{x}$$
The cost function in each iteration are chosen at random and uniformly from $\{f_v | v \in V \}$. So in iteration $t$ first  algorithm $\Acal$  chooses $x_t$ and we choose random $v_t$ and incur cost function $f_t(x) = \inner{v_t}{x}$. Now we want to compute $\EE(\Rcal_T)$.
\[
\begin{aligned}
	\EE(\Rcal_T) = ~ & \EE\left(\sum\limits_{t=1}^T f_t(x_t) - \min\limits_{x \in \Kcal} \sum\limits_{t=1}^T f_t(x) \right)\\
	= ~& \sum\limits_{t=1}^T \EE(\inner{v_t}{x_t}) - \EE\left( \min\limits_{x \in \Kcal} \sum\limits_{t=1}^T \inner{v_t}{x}\right) \\
	\stackrel{\text{(a)}}{=} ~& \sum\limits_{t=1}^T \inner{\EE(v_t)}{\EE(x_t)}) - \EE\left( \min\limits_{x \in \Kcal} \inner{\sum\limits_{t=1}^T v_t}{x}\right)\\
	\stackrel{\text{(b)}}{=} ~&  - \EE\left( \min\limits_{x \in \Kcal} \inner{\sum\limits_{t=1}^T v_t}{x}\right)\\
\end{aligned}
\]
where (a) follows by the fact that $\{v_t\}_{t \in [T]}$ are i.i.d. and $x_t$ is depends on just $v_{1:t-1}$ so $x_t, v_t$ are independent, (b) is due to $\EE(v_t) = 0$. Now suppose that $S_T = \sum\limits_{t=1}^T v_t$ so we should compute $\EE( \min\limits_{x \in \Kcal} \inner{S_T}{x})$. Since $\Kcal$ is symmetric with respect to the origin, so for every vector $y \in \RR^d$ we have $\max\limits_{x \in \Kcal} \inner{y}{x} =  -\min\limits_{x \in \Kcal} \inner{y}{x}$, as a consequence we should  calculate $\EE(\max\limits_{x \in \Kcal} \inner{y}{x})$. 
On the other hand we know that:
 \[\max\limits_{x \in \Kcal} \inner{S_T}{x} = \frac{D}{2}\max\limits_{\{x | \|x\| \leq 1\}} \inner{S_T}{x} = \frac{D}{2}\|S_T\|_.^*
 \]
  So 
  \begin{equation}\label{eq:s1}
  	\EE(\Rcal_T) =   \frac{D}{2} \EE(\|S_T\|_.^* ) 
  \end{equation}
Now we want to give a lower bound for $\|S_T\|_.^*$. Now we know that 
 By the Cauchy-Schwarz inequality we can say that 
 \[
 \begin{aligned}
 \sum\limits_{i=1}^{d} a_i |S_T(i)| & \leq \|S_T\|_.^* \|\bm{e}\|_.\\
 & \leq \|S_T\|_.^* \sum_{j=1}^d \|\bm{e}_i\|_. \\
 & = d \|S_T\|_.^*
 \end{aligned}
 \]
  where $\bm{e}(i)  = sign(S_T(i))a_i $. So by using \eqref{eq:s1} we have:
 \begin{equation}\label{eq:s2}
 \frac{D}{2d}\sum_{i=1}^{d} a_i \EE(|S_T(i)|) \leq \EE(\Rcal_T)	
 \end{equation}
 Now we know that $S_T(i) = \sum\limits_{t = 1}^T v_t(i)$
 and by considering i.i.d.  property of $\{v_t\}_{t \in [T]}$, using central limit theorem result in $S_T(i) \simeq N(0, T\sigma^2) $ where $\sigma^2 = var(v_t(i))$ that is simply $\frac{L^2 a_i^2}{d}$. Now it is sufficient to compute $\EE(|S_T(i)|)$.
 \[
 \begin{aligned}
 	\EE(|S_T(i)|) \simeq &\frac{1}{\sqrt{2T\pi}\sigma} \int_{-\infty}^{\infty} |x| e^{\frac{-x^2}{2T\sigma^2}} dx\\
 	= & \frac{\sqrt{2}}{\sqrt{T\pi}\sigma} \int_{0}^{\infty} x e^{\frac{-x^2}{2T\sigma^2}} dx\\
 	= & \frac{-\sqrt{2T}\sigma e^{\frac{-x^2}{2T\sigma^2}}}{\sqrt{\pi}} |_0^\infty \\
 = &  	\frac{\sqrt{2T}L a_i}{\sqrt{d\pi}}
 \end{aligned}
 \] 
 So $\frac{D}{2d}\sum\limits_{i=1}^d a_i \EE(|S_T(i)|) \simeq  \frac{D}{2d}(\sum\limits_{i=1}^d a_i^2)\frac{\sqrt{2T}L}{\sqrt{d\pi}}$ .
 Now using result \eqref{eq:s2} leads to having following bound:
 \[
\EE(\Rcal_T) = \Omega(LD\sqrt{T})
 \]
 This result show that there are sample vectors $\{\hat{v}_t\}_{t \in [T]}$ that regret of cost functions $\{f_{\hat{v}_t}\}_{t \in [T]}$  incurs $\Omega(LD\sqrt{T})$, anyway. 
\end{proof}

%
\begin{proof}[Proof of Corollary 3.3]
Let $R_T$ be the random variable where obtained by Framework \ref{alg:framework}. Suppose $p_t$ be the probabilities over expert in round $t$. Then for regret bound of modified version of the framework we have:
\[
\begin{aligned}
\Rcal_T & = \sum\limits_{t=1}^T f_t(x_t) -  \min\limits_{x \in \Dcal}\sum\limits_{t=1}^T  f_t(x)\\
& = 	\sum\limits_{t=1}^T f_t\left(\sum\limits_{i=1}^K x_t^i p_t(i)\right) -  \sum\limits_{t=1}^T \min\limits_{x \in \Dcal} f_t(x)\\
& \stackrel{(a)}{\leq} 
\sum\limits_{t=1}^T \sum\limits_{i=1}^K p_t(i)f_t(x_t^i) - \min\limits_{x \in \Dcal}\sum\limits_{t=1}^T f_t(x)\\
& = \sum\limits_{t=1}^T \EE(f_t(x_t^i)) -  \min\limits_{x \in \Dcal}\sum\limits_{t=1}^T f_t(x)\\
& = \EE(R_T)
\end{aligned}
\]
where (a) follows by Jensen's inequality.
\end{proof}

\subsection{Proof of Theorem 3.6}
\begin{proof}{Proof of Theorem 3.6}
Let $x^* = \arg \min\limits_{x \in \Dcal} \sum\limits_{t=1}^T \inner{\grad_t}{x}$ and $a$ is arbitrary optimizer in $[K]$. For the regret of this algorithm we can write:
\begin{equation}\label{eq:s1}
\begin{aligned}
\Rcal_T & \leq \sum\limits_{t=1}^T f_t(x_t) -\sum\limits_{t=1}^T f_t(x^*) \\
& \stackrel{\text{(a)}}{\leq} \inner{\grad_t}{x_t-x^*}\\
& \stackrel{\text{(b)}}{=} \sum\limits_{t=1}^T \inner{\grad_t}{\sum\limits_{i=1}^K x_t^i p_t(i)} - 
\sum\limits_{t=1}^T \inner{\grad_t}{x_t^{a}}  +
 \sum\limits_{t=1}^T \inner{\grad_t}{x_t^{a}} -
 \sum\limits_{t=1}^T \inner{\grad_t}{x^*} \\
& = 2F\left(\sum\limits_{t=1}^T (\inner{p_t}{\ell_t} - \ell_t(a))\right) +  \left(\sum\limits_{t=1}^T \widehat{f}_t(x_t^{a}) - \sum\limits_{t=1}^T \widehat{f}_t(x^*)\right)\\
& = 2F\left(\sum\limits_{t=1}^T \EE(\ell_t - \ell_t(a))\right) +  \left(\sum\limits_{t=1}^T \widehat{f}_t(x_t^{a}) - \sum\limits_{t=1}^T \widehat{f}_t(x^*)\right)\\
& = 2F \EE(\Rcal_T^{\Acal}) + \Rcal_T^{\OCO_{a}} 
\end{aligned}
\end{equation}
where (a) follows by convexity of $\{f_t\}_{t \in [T]}$ and (b) follows by the fact that $x_t = \sum\limits_{i=1}^K x_t^i p_t(i)$.\\
By the Theorem \ref{theorem:squint} we have bound for $\EE(\Rcal_T^\Acal) \leq 2\sqrt{V_T(i)\ln k}$. So we can rewrite \eqref{eq:s2} as following:
\begin{equation}\label{eq:8}
\begin{aligned}
\Rcal_T &\leq 4F\sqrt{V_T(i)\ln K} + \Rcal_T^{\OCO_i} \Rightarrow \\ 
& \Rcal_T \leq \min\limits_{i \in [K]} \left( 4F\sqrt{V_T(i)\ln K} + \Rcal_T^{\OCO_i} \right)
\end{aligned}
\end{equation}
Suppose $a^* = \arg \min\limits_{i \in [K]} \Bcal_T^{\OCO_i}$. By the assumption 3 that we had in \S~\ref{sec:preliminaries} this algorithm $\OCO_{a^*}$ should perform on  a family of $L$-Lipschitz cost functions and domains with diameter $D$, both w.r.t. some norm $\|.\|$. So by using Lemma \ref{lemma:lowerbound} we can say that
\begin{equation}\label{eq:6}
\begin{aligned}
		 \Bcal_T^{\OCO_{a^*}} &= \Omega(LD\sqrt{T}) \Rightarrow\\
	 \sqrt{\ln K}\Bcal_T^{\OCO_{a^*}} &= \Omega(\sqrt{T\ln K}LD) 
	 \end{aligned}
\end{equation}
Using Lemma \ref{lemma:l2} result in $F \leq LD$. Also we know that $V_T(i) = \sum\limits_{t =1}^{T}  (\inner{p_t}{\ell_t} - \ell_t(i))^2$ and according to the fact that $\forall i : \ell_t(i) < 1$ then we can bound $V_T(i)$ hence : $V_T(i) \leq T$. So we have:
\begin{equation}\label{eq:7}
	4F\sqrt{V_T(i)\ln K} \leq 4\sqrt{\ln K} LD\sqrt{T} = 
	\Ocal(\sqrt{\ln K} LD\sqrt{T})
\end{equation}

 Now using \eqref{eq:6} and \eqref{eq:7}, result in $4F\sqrt{V_T(i)\ln K} = \Ocal\left(\sqrt{\ln K}\Bcal_T^{\OCO_{a^*}}\right)$ and  using Equation \eqref{eq:8} leads to $\Rcal_T =  \Ocal(\sqrt{\ln K}\Bcal_T^{a^*})$.
\end{proof}

\subsection{Proof of Theorem 3.10.}
\begin{proof}[Proof of Theorem 3.10]
	Let $x^* = \arg \min\limits_{x \in \Dcal} \sum\limits_{t=1}^T \inner{\grad_t}{x}$ and $i$ is arbitrary optimizer in $[K]$. As we mentioned in Proposition \ref{prop:framework} for the regret of this algorithm we can write:
\begin{equation}\label{eq:s1}
\Rcal_T \leq 2F \Rcal_T^{\Acal} + \Rcal_T^{\OCO_{i}} 
\end{equation}
We have following upper bound for regret of $\OCO_i$.
 $$\Rcal_T^{\OCO_i} \leq \Bcal_T^{\OCO_i}$$
So we can say that:
$$\EE(\Rcal_T) \leq 2F \EE(\Rcal_T^\Acal) + \Bcal_T^{\OCO_i}$$
On the other hand, from Theorem \ref{thm:GBPA_RegretBnd}, Corollary \ref{corollary:exp3} and Lemma \ref{lemma:l1} we have following regret bound for algorithm $\Acal$:
\begin{equation}\label{eq:9}
\begin{aligned}
	&\alpha = 1/2\  (GBPA):   \EE(\Rcal_T) \leq  8F\sqrt{T  K},\\
		&\alpha \rightarrow 1 \ (EXP3): \EE(\Rcal_T) \leq  4F\sqrt{T  K \ln K}
\end{aligned}
\end{equation}

By the assumption 3 that we had in \S~\ref{sec:preliminaries} this algorithm $\OCO_{i}$ should perform on  a family of $L$-Lipschitz cost functions and domains with diameter $D$, both w.r.t. some norm $\|.\|$.
 So by using Lemma \ref{lemma:lowerbound} we can say that
\begin{equation}\label{eq:10}
\begin{aligned}
		 \Bcal_T^{\OCO_{i}} &= \Omega(LD\sqrt{T}) \\
\Rightarrow	 \sqrt{K}\Bcal_T^{\OCO_{i}} &= \Omega(\sqrt{TK}LD),\\
	 \sqrt{K\ln K}\Bcal_T^{\OCO_{i}} &= \Omega(\sqrt{TK\ln K}LD) 
	 \end{aligned}
\end{equation}
Using Lemma \ref{lemma:l2} result in $F \leq LD$. Combining \eqref{eq:9}, \eqref{eq:10} and the fact that $F \leq LD$,  result in :
\[
\begin{aligned}
	&\alpha = 1/2\  (GBPA):   \EE(\Rcal_T) \leq  \Ocal(\sqrt{K} \Bcal_T^{\OCO_i}),\\
		&\alpha \rightarrow 1 \ (EXP3): \EE(\Rcal_T) \leq  \Ocal(\sqrt{K\ln K} \Bcal_T^{\OCO_i})
\end{aligned}
\]
\end{proof}

\subsection{Proof of Theorem 3.13 }
\begin{proof}
According to Theorems \ref{theorem:master} and \ref{theorem:Fast_master} we have following bound for MGD and FMGD on the mentioned setting.
\begin{equation}\label{eq:11}
\begin{aligned}
	\text{MGD:} ~~~  &\Rcal_T \leq 4F\sqrt{T \ln K} + \Bcal_T^{\OCO_i} ,\\
	\text{FMGD:}~~~ & \EE(\Rcal_T) \leq 8F\sqrt{T K } + \Bcal_T^{\OCO_i} 
\end{aligned}
\end{equation}
Suppose that $d_i$ be diameter of $\Dcal$ w.r.t. norm $\|.\|_i$. By strongly convexity of $\varphi_i$ we know that
\begin{equation}\label{eq:12}
\begin{aligned}
& B_{\varphi_i}(x,y)  \geq \frac{1}{2} \rho_i \|x-y\|_i^2  \\
&\Rightarrow   \sup\limits_{x,y \in \Dcal} B_{\varphi_i}(x,y) \geq \sup\limits_{x,y \in \Dcal}  \frac{1}{2}\rho_i \|x-y\|_i^2 \\
&\Rightarrow  D_i\geq \frac{1}{2} \rho_i d_i^2
\end{aligned}
\end{equation}
Now by Lemma \ref{lemma:l1} we have $F \leq L_i d_i$. So by using \eqref{eq:12} we can see that: 
$F \leq L_i \sqrt{\frac{2 D_i }{\rho_i}}$. \\
According to the fact that $\OCO_i$ is a OMD algorithm then by using  \ref{theorem:omd} leads to $\Bcal_T^{\OCO_i} = L_i \sqrt{\frac{2D_i T}{\rho_i}}$ so by using these results and combining with \eqref{eq:11} we have
\[
\begin{aligned}
	\text{MGD:} ~~~  &\Rcal_T \leq (4\sqrt{\ln K} + 1)L_i \sqrt{\frac{2D_i T}{\rho_i}} ,\\
	\text{FMGD:}~~~ & \EE(\Rcal_T) \leq (8\sqrt{K} + 1)L_i \sqrt{\frac{2D_i T}{\rho_i}}
\end{aligned}
\]
\end{proof}

		\section{Domain Specific Example}\label{appendix:domain}
The goal of this section is to examine the intrinsic difference between \emph{Quadratic}  and \emph{Negative Entropy} regularizer, when the optimization domain is a $\simplex(d)$ and $\BB_2$ Ball. Our goal in learning the best regularizer among family of regularizers, achieved by providing two main algorithm. We experimented these algorithms on two domains $\BB_2$-Ball and $\simplex(d)$. In the following we want to find the best regularizer with two choices of regularizer for these domains, that can verify our experimental results on proposed algorithms.  
 We study all four possible combinations in the following sections. Finally we will propose an regularizer function called \emph{Hypentropy} that has parameter $\beta$ that tuning it leads to covering both Negative Entropy and  Quadratic. 
 
\subsection{Computing Bregman Divergence} 
Negative Entropy is:
$$R(x) = \sum\limits_{i=1}^d x_i \log x_i$$
  Quadratic is:
$$R(x) = \frac{1}{2}\|x\|_2^2$$
For Quadratic we have
 $$B_R(x,y)=\frac{1}{2}\left \| x-y \right \|_2^2$$  
and for Neg Entropy we have
 $$B_R(x,y)=\sum_{i=1}^{d}y_i-\sum_{i=1}^{d}x_i + \sum_{i=1}^{d}x_i\log\frac{x_i}{y_i}$$
  
so we are going to  compare these regularizers on two domain : $\simplex(d)$ and $\BB_2$.

\subsection{Quadratic Regularisation on $\BB_2$}

Let $R(x)=\frac{1}{2}\left \| x \right \|_2^2$ and $K=\left \{ x:\left \| x-x_0 \right \|_2\leq 1 \right \}$ according to the definition of mirror descent algorithm we have:
\[y_{t+1} = x_t-\eta _t\nabla f(x_t)\]
\[x_{t+1} =\proj{K}{R}(x_t-\eta _t\nabla f(x_t))=\frac{x_t-\eta _t\nabla f(x_t)-x_0}{\left \| x_t-\eta _t\nabla f(x_t) - x_0 \right \|_2}+x_0\]
\textbf{Analysis:}
using \eqref{eq:2} we have the following bound:
\[
 \Rcal_T \leq \Ocal(L_{f,2} \sqrt{2T} ) 
 \]

\subsection{Entropic Regularization on $\BB_2$}
given that $R(x)=-\sum_{i=1}^{d}x_ilog(x_i)$ and $K=\left \{ x:\left \| x-x_0 \right \|_2\leq 1 \right \}$ the projection using this norm seems not to have a simple analytical soloution and we can use numerical methods such as gradient descent.
\\

\textbf{Analysis:}
 We know that $R(x)=-\sum_{i=1}^{d} |x_i| \log(|x_i|)$ is 1-strongly convex w.r.t. $\left \| . \right \|_1$.  If we pick $x_0 = \sqrt{d}^{-1}\bm{1}$ as a start point, to obtain the upper bound for regret we need to just calculate $D_R = \sup_{x \in K}B_R(x,x_0)$ and  bound $\left \|. \right \|_{\infty}$ with $\left \|. \right \|_{2}$ in order to compare with Quadratic regularizer regret bound.\\
 First we provide bellow lemma:
 \begin{lemma}\label{lemma:ent}
if $z_i \geq 0$ and $\sum_{i=1}^d z_i^2 \leq c^2$ then 
$ 	\sum_{i=1}^d z_i \log(z_i) \leq c^2$ 
\begin{proof}
It's sufficient to consider the following inequality:
$$\forall z \in \RR^+ : \log z \leq z-1 < z $$
\end{proof}
\end{lemma}

Assume that $y_i = |x_i|$ so we have the following bound on $D_R$ : 
\begin{equation*}\label{eq:pareto mle2}
\begin{aligned}
	D_R = \underset{y \in K}{\sup}B_R(y,x_0)= &
	\sum_{i=1}^d y_i \log{y_i} + \sqrt{d} \log{\sqrt{d}} + \sum_{i=1}^d (\log{\sqrt{d}}-1)(y_i- \frac{1}{\sqrt{d}})\\
	= & \sum_{i=1}^d y_i \log{\frac{\sqrt{d}}{e}y_i}  + \sqrt{d} (z_i := \frac{\sqrt{d}}{e}y_i, c := \frac{\sqrt{d}}{e}) \\
	= & \frac{\sum_{i=1}^d z_i \log{z_i}}{c}  + \sqrt{d} \hspace{2.5pt} \text{(Applying lemma  \ref{lemma:ent})}\\
	\leq & c + \sqrt{d} =\frac{\sqrt{d}}{e} +\sqrt{d}= O(\sqrt{d})\\
\end{aligned}
\end{equation*}

For $\|. \|_{\infty}$ its easy to check that  $\forall x \in \RR^d $ : 
\begin{equation}
\frac{\|x\|_{2}}{\sqrt{d}} \leq \|x\|_{\infty} \leq \|x\|_{2}
\end{equation}\label{eq:ineq}
hence using equation (\ref{eq:2}) we have the following bound:
\begin{equation}
\Rcal_T \leq O( L_{f,\infty} \sqrt{2D_R T}) 
\end{equation}
\\

\begin{corollary}
If  our domain is unit ball ($\BB_2$) then using quadratic regularizer gives us better regret bound comparing to negative entropy. To be more precise if we assume that upper bound for regret with respect to negative entropy and quadratic on $\BB_2$ are $\Bcal_T^{Ent}$ and $\Bcal_T^{Quad}$, respectively, then we have the following inequality :
\begin{equation}
1 \leq \frac{\Bcal_T^{Ent}}{\Bcal_T^{Quad}}  \leq \sqrt{d}	
\end{equation}
\end{corollary}

\subsection{Quadratic Regularisation on $\simplex(d)$}
given that $R(x)=\frac{1}{2}\left \| x \right \|_2^2$ and \[K = \left \{x=(x_1,x_2,...,x_d)\in \mathbb{R}^d:\sum_{i=1}^{d}x_i=1,\forall i, 1\leq i\leq d  \right \}\]
first note that the projection onto the probability simplex using euclidean norm is very easy using KKT and you can see the following algorithm.

\begin{algorithm}[H]
\caption*{Euclidean projection of a vector onto the probability simplex}
\begin{algorithmic}
    \STATE {\textbf{Input:} $y \in \mathbb{R}^d$}
        \STATE{Sort \textbf{y} into \textbf{u}:$u_1 \geq u_2 \geq ... \geq u_d$}
        \STATE Find \textbf{$\rho$} = $\max\left \{ 1\leq j\leq d:u_j + \frac{1}{j}(1-\sum_{i=1}^{j}u_i) > 0 \right \}$
        \STATE {Define \textbf{$\lambda$}=$\frac{1}{\rho}(1-\sum_{i=1}^{\rho}u_i) > 0$} 
    \STATE{ \textbf{Output:} output x, s.t. $x_i = \max\left \{ y_i+\lambda,0 \right \}$} 
\end{algorithmic}
\end{algorithm}
 hence we have:
\[y_{t+1} = x_t-\eta _t\nabla f(x_t)\]
\[x_{t+1} =\proj{K}{R}(x_t-\eta _t\nabla f(x_t))\]
and the projection is as we have defined above.

\textbf{Analysis:}
It is easy to check that $D_R$ is constant. So by using (\ref{eq:2}) we have the following bound:

\begin{equation}\label{eq:4} \Rcal_T \leq O(L_{f,2}\sqrt{2T}) 
\end{equation}

\subsection{Entropic Regularisation on $\simplex(d)$}

given that $R(x)=-\sum_{i=1}^{d}x_i \log(x_i)$ and \[K = \left \{x=(x_1,x_2,...,x_d)\in \mathbb{R}^d:\sum_{i=1}^{d}x_i=1,\forall i, 1\leq i\leq d  \right \}\]
we can easily show that projection onto the $\simplex(d)$ using Bregman Divergence is like normalizing using L1-norm. Hence by the update rule of OMD we have
\[-1+\log(y_{t+1}^{(i)}) = -1+\log(x_{t}^{(i)})-\eta _t\nabla f(x_t^{(i)})\]
so:
\[y_{t+1}^{(i)} = x_{t}^{(i)}\exp(-\eta _t\nabla f(x_t^{(i)}))\]
\[x_{t+1}^{(i)} =\proj{K}{R}(x_{t}^{(i)}\exp(-\eta _t\nabla f(x_t^{(i)}))) = \frac{x_{t}^{(i)}\exp(-\eta _t\nabla f(x_t^{(i)}))}{\sum_{i=1}^{d}x_{t}^{(i)}\exp(-\eta _t\nabla f(x_t^{(i)}))}\]
\textbf{Analysis:}
pick $x_0 = d^{-1}1$ and $R(x)=-\sum_{i=1}^{d}x_i \log(x_i)$ is 1-strongly convex w.r.t. $\left \| . \right \|_1$ . Then 
\[\underset{x \in K}{\sup}B_R(x,x_0)=\underset{x \in K}{\sup}KL(x,x_0)=\log(d) + \underset{x \in K}{\sup}\sum_{i=1}^{d}x_i \log(x_i) \leq \log(d)\]
hence using (\ref{eq:2}) we have the following bound:
\begin{equation}\label{eq:5}
	 \Rcal_T \leq \Ocal( L_{f,\infty} \sqrt{2\log(d)T})  
\end{equation}

\begin{corollary}
	If our domain is $\simplex(d)$ then based on equations \ref{eq:ineq}, \ref{eq:4} and \ref{eq:5} we can compare regret bound of two regularizers (Quadratic and negative entropy) on this domain :
	\begin{equation}
\frac{1}{\sqrt{\log d}} \leq \frac{\Bcal_T^{Quad}}{\Bcal_{T}^{Ent}}  \leq \sqrt{\frac{d}{\log d}}	
\end{equation}
\end{corollary}

So we can say that based on Lipschitzness of our functions sometimes negative entropy has better performance than quadratic and sometimes vise versa. But our intuition tell us that equation \ref{eq:ineq} on average is near the lower bound instead of upper bound and as consequence the lower bound of above corollary can be substitute with 1 which result in better performance of negative entropy than quadratic on $\simplex(d)$. 

\subsection{Hypentropy}

Here we introduce a regularizer that covers both negative entropy and quadratic norm.\\

\begin{definition} (Hyperbolic-Entropy)
	For all $\beta > 0$, let $\phi_\beta\colon \RR^d \to \RR$ be defined as:
	\begin{equation}
	\phi_\beta(x)=\sum_{i=1}^d x_i \arcsin(\frac{x_i}{\beta}) - \sqrt{x_i^2 + \beta^2}
\end{equation}
\end{definition}

The Bregman divergence is a measure of distance between two points defines in term of strictly convex function and here for hypentropy function $\phi_\beta$ is driven like below:
\begin{equation}
B_\phi(x , y) = \phi_\beta(x) - \phi_\beta(y) - \langle \nabla \phi_\beta(x), (x-y) \rangle
\end{equation}
\[= \sum_{i=1}^d [ x_i ( \arcsin(\frac{x_i}{\beta}) - \arcsin(\frac{y_i}{\beta})) - \sqrt{x_i^2 + \beta^2} + \sqrt{y_i^2 + \beta^2} ] \]
	
The key reason that makes this function behave like both euclidean distance and relative entropy is that the hessian of hypentropy would be interpolate between hessian of both functions while vary parameter $\beta$ from 0 to $\infty$.

First we calculate the hessian of $\phi_\beta$ to compare it to Euclidean distance and entropy function:
\[\phi_\beta''(x) = \frac{1}{\sqrt{x^2+\beta^2}}\]
Now consider  $x\ll\beta$ then $\phi_\beta''(x) = \frac{1}{\beta}$ is constant function similar to euclidean distance and in other case consider $\beta=0$ then $\phi_\beta''(x) = \frac{1}{|x|}$ and it is the same as hessian of negative entropy.

\subsubsection{Diameter calculations for hypentropy}

In this section we calculate the diameter for both $\simplex(d)$ and $\BB_2$

First we need good approximation for bregman divergence of hypentropy function:
\[B_\phi(x , 0) = \phi_\beta(x) - \phi_\beta(0)\]
\[=\sum_{i=1}^d (x_i \arcsin(\frac{x_i}{\beta}) - \sqrt{x_i^2 + \beta^2}) + \beta d\]
\[\leq\sum_{i=1}^d x_i \arcsin(\frac{x_i}{\beta})\]
\[=\sum_{i=1}^d |x_i|\log(\frac{1}{\beta}(\sqrt{x_i^2 + \beta^2} + |x_i|)) \]
Thus WLOG,
\[ \mathrm{diam}(\BB_2) \leq \sum_{i=1}^d x_i \log(\frac{1}{\beta}(\sqrt{x_i^2 + \beta^2} + x_i )) \]
\[ \leq \sum_{i=1}^d x_i \log (1 + \frac{2x_i}{\beta}) \]
\[ \leq \sum_{i=1}^d \frac{2x_i^2}{\beta} = \frac{2\|x\|_2^2}{\beta} \leq \frac{2}{\beta} \]
For $\simplex(d)$ if consider $\beta \leq 1$ this inequality holds, $\sqrt{x_i^2 + \beta^2} + x_i \leq \sqrt{2} + 1 < 3$. thus, it is clear that:
\[ \mathrm{diam}(\simplex(d)) < \sum_{i=1}^d x_i \log (\frac{3}{\beta}) = \|x\|_1 \log(\frac{3}{\beta}) = \log(\frac{3}{\beta}) \]
and if $\beta \geq 1$ also we have, $\sqrt{x_i^2 + \beta^2} + x_i \leq \beta \sqrt{1 + (\frac{x_i}{\beta})^2} + x_i \leq \sqrt{2}\beta + 1$. Hence, in this case we have:
\[ \mathrm{diam}(\simplex(d)) \leq \sum_{i=1}^d x_i \log (\frac{\sqrt{2}\beta + 1}{\beta}) \leq  \sum_{i=1}^d x_i \log(\sqrt{2} + 1) = \|x\|_1 \log(\sqrt{2} + 1) < \log(3) \]

	\end{document}